\documentclass[preprint,12pt]{elsarticle}

\usepackage{amsmath, amsfonts, amssymb}
\usepackage{mathtools}
\mathtoolsset{showonlyrefs}
\usepackage{dsfont}
\usepackage{bm}
\usepackage[dvipsnames]{xcolor}
\usepackage{amsthm}

\newtheorem{thm}{Theorem}
\newtheorem{rem}{Remark}

\usepackage{algorithm}
\usepackage[noend]{algorithmic}

\usepackage{array}
\usepackage[caption=false,font=normalsize,labelfont=sf,textfont=sf]{subfig}
\usepackage{textcomp}
\usepackage{stfloats}
\usepackage{url}
\usepackage{verbatim}
\usepackage{graphicx}

\usepackage{yaacro}
\def\replycolor{black!30!blue}
\usepackage[font=footnotesize]{caption}
\newcommand{\rev}[1]{{#1}}

\usepackage{hyperref}
\hypersetup{
    colorlinks=true,
    linkcolor=\replycolor,
    filecolor=\replycolor,      
    urlcolor=\replycolor,
    citecolor = \replycolor,
    pdftitle={A Semi-Lagrangian Approach for Time and Energy Path Planning Optimization in Static Flow Fields},
}

\begin{acgroupdef}[list=acronimos]
    \acdef{CPI}{Concurrent Policy Iteration}
    \acdef{MEPI}{Multi-objective Evolutionary Policy Iteration}
    \acdef{HJB}{Hamilton-Jacobi-Bellman}
    \acdef{SL}{semi-Lagrangian}
    \acdef{DP}{Dynamic Programming}
    \acdef{CNPq}{Conselho Nacional de Desenvolvimento Científico e Tecnológico}
    \acdef{CAPES}{Coordenação de Aperfeiçoamento de Pessoal de Nível Superior}
    \acdef{FAPEMIG}{Fundação de Amparo à Pesquisa do Estado de Minas Gerais}

    \acdef{OP}{Orienteering Problem}
    \acdef{WSN}[WSN]{Wireless Sensor Network}
\end{acgroupdef}

\journal{Journal of the Franklin Institute}

\begin{document}

\begin{frontmatter}



\title{A Semi-Lagrangian Approach for Time and Energy Path Planning Optimization in Static Flow Fields}



\author[delt]{Víctor C. da S. Campos}
\author[delt]{Armando A. Neto}
\author[dcc]{Douglas G. Macharet}

\affiliation[delt]{organization={Dept. of Electronics Engineering, Universidade Federal de Minas Gerais},
            addressline={Av. Antônio Carlos, 6627, Pampulha}, 
            city={Belo Horizonte},
            postcode={31270-901}, 
            state={Minas Gerais},
            country={Brazil}}

\affiliation[dcc]{organization={Computer Vision and Robotics Laboratory (VeRLab), Dept. of Computer Science, Universidade Federal de Minas Gerais},
            addressline={Av. Antônio Carlos, 6627, Pampulha}, 
            city={Belo Horizonte},
            postcode={31270-901}, 
            state={Minas Gerais},
            country={Brazil}}

\begin{abstract}
Efficient path planning for autonomous mobile robots is a critical problem across numerous domains, where optimizing both \emph{time} and \emph{energy} consumption is paramount. 
This paper introduces a novel methodology that considers the dynamic influence of an environmental flow field and geometric constraints, including obstacles and forbidden zones, enriching the complexity of the planning problem. 
Here, we formulate it as a multi-objective optimal control problem, and propose a novel transformation called \emph{Harmonic Transformation}, applying a semi-Lagrangian scheme to solve it.
\rev{The set of Pareto efficient solutions is obtained considering two distinct approaches: $i$) a deterministic method referred to as \ac{CPI}; and $ii$) an evolutionary-based one, called \ac{MEPI}. Both methods were designed to make use of the proposed Harmonic Transformation.}
\rev{Through an extensive analysis of these approaches, comparing them with the state-of-the-art literature, we demonstrate their efficacy in finding optimized paths.}
\rev{Generally speaking, the Pareto Set of solutions found in our experiments indicates that the \ac{CPI} demonstrated better performance in finding solutions close to the time-optimal one, whereas the \ac{MEPI} was most successful in finding solutions close to the energy-optimal solution.}
\end{abstract}




\end{frontmatter}



\newcommand{\escalar}[1]{\ensuremath{\mathit{#1}}}
\newcommand{\vetor}[1]{\ensuremath{\boldsymbol{#1}}}
\newcommand{\matriz}[1]{\ensuremath{\mathbf{\uppercase{#1}}}}
\newcommand{\conjunto}[1]{\ensuremath{\mathcal{\uppercase{#1}}}}
\newcommand{\distribuicao}[1]{\ensuremath{\mathcal{\uppercase{#1}}}}
\newcommand{\transpose}{\ensuremath{{}^\intercal}}
\newcommand{\espaco}[1]{\ensuremath{\mathds{\MakeUppercase#1}}}

\newcommand{\xv}{\mathbf{x}}
\newcommand{\uv}{\mathbf{u}}
\newcommand{\yv}{\mathbf{y}}

\newcommand{\Uset}{\conjunto{U}}

\newcommand{\Obstacles}{\conjunto{O}}

\newcommand{\reais}{\espaco{R}}
\section{Introduction}


The continual research and development of new and more advanced path-planning approaches play a pivotal role in Robotics~\cite{LaValle2006Planning}. Such techniques enable autonomous mobile robots to navigate efficiently and safely in complex and dynamic environments, making them essential for diverse applications, from logistics to monitoring and exploration.
In this context, new challenges arise when robotic systems address singular and multiple objectives and often conflicting goals. These objectives range from minimizing \emph{travel time} and \emph{energy consumption} to optimizing factors like safety and resource allocation~\cite{Hohmann2021Hybrid}.
Furthermore, it is also imperative to acknowledge that, in several domains, environmental dynamics substantially influence the trajectories and behaviors of the vehicles. This is particularly evident in fields such as aerospace, where factors like air density, wind patterns, and gravitational forces intricately shape the aircraft flight paths~\cite{Cole2018Reactive}. 

Similarly, in maritime environments, the varying properties of water, including currents and turbulence, substantially impact the maneuverability of underwater or surface vehicles~\cite{Weekly2014Autonomous}. 
Hence, an adequate navigation strategy holds the potential to generate paths that optimize robot movement according to the surrounding flow field resulting from atmospheric and ocean currents. This synergy between path planning and environmental dynamics enhances the efficiency and speed of vehicle navigation and bolsters adaptability, ensuring that robots can navigate seamlessly through environments where flow dynamics are significant. Ultimately, this approach fosters improved resource utilization and reduced energy consumption, increasing the system's performance across a broad spectrum of robotic applications.

In this paper, we introduce an innovative approach to deal with time and energy-efficient path planning within environments characterized by static flow fields and stationary obstacles. This problem poses a multi-objective optimal control problem with forbidden zones in the state space, and we solve it with the following contributions:
\begin{itemize}
    \item A \emph{novel} transformation, called \emph{Harmonic Transformation}, is employed to map values onto the $[0, 1]$ range to deal with the forbidden zones and avoid possible numerical problems in the computation of the value functions. We show that, given that the corresponding assumptions hold, a semi-Lagrangian approach converges to the unique viscosity solution of the corresponding transformed partial differential equations;
    \item Considering that time and energy are usually conflicting costs, we propose two approaches to find the set of \emph{Pareto efficient solutions} in a multi-objective manner: a \emph{deterministic} one, based on solving multiple single-objective optimizations concurrently; and a \emph{metaheuristic} approach, based on a proper multi-objective evolutionary algorithm.
\end{itemize}

\rev{Semi-Lagrangian methods \cite{Falcone2013} are advantageous for path planning in real-time, dynamic environments due to their stability with larger time steps and computational efficiency. Their formulation allows for integrating dynamic constraints, like motion and energy limits, into cost functions, enhancing adaptability to real-world scenarios. Their support for unstructured and adaptive grids, suitable for high-dimensional and/or cluttered spaces, is resource-efficient and ideal for complex, real-world applications such as robotic navigation and obstacle avoidance. In addition, the fact that they find a closed-loop policy for the planning problem ensures greater robustness whenever they are employed on real systems.}

The remainder of this paper is structured as follows: 
\rev{Section \ref{sec:related_work} presents the state-of-art related work;}
Section \ref{sec:problem} delineates the problem formulation and introduces the Harmonic Transformation, demonstrating its application within a dynamic programming framework;
Section \ref{sec:proposed_approaches} elaborates on the two distinct approaches we propose to address the multi-objective problem;
Section \ref{sec:numSim} presents numerical results, showcasing the efficacy of our proposed approaches;
Lastly, in Section \ref{sec:conc}, we provide concluding remarks and outline potential avenues for future research.

\rev{
\section{Related work}
\label{sec:related_work}


In single-objective path planning approaches, the most commonly prioritized factors are path length~\cite{Zafar2018Methodology,Low2019Solving} and travel time~\cite{Gasparetto2015Path,Foehn2021Timeoptimal}. However, enhancing solution quality and applicability can often be achieved by incorporating additional attributes, such as path safety or vulnerability, and smoothness~\cite{Ahmed2013Multiobjective,Xue2018Solving}.


However, problems are complex in the real world, and multi-objective formulations have emerged as a noteworthy approach to providing solutions in challenging scenarios. For instance, \cite{MA2018Multiobjective} introduces a particle swarm optimization-based algorithm that considers multiple objectives, including travel length, path smoothness, economic cost, and path safety. 
%
Regarding the more general class of routing problems, where a sequence of visits is required, a multi-objective version of the \ac{OP} was introduced in~\cite{Macharet2021Minimal}, aiming to maximize cumulative reward while simultaneously minimizing exposure to sensors deployed in the environment. The \ac{OP} has also been studied in environments with flows~\cite{Mansfield2022Energy,Mansfield2023Energy}. In this multi-objective formulation, the aim is not only to maximize the collected reward but also to minimize energy expenditure by utilizing the surrounding environmental dynamics.
In another recent study, \cite{Doshi2023EnergyTime} employs a level set method to ascertain energy-time optimal solutions within dynamic flow environments.

The authors of \cite{zhang2024many} introduce the DSFMO algorithm, a multi-objective evolutionary approach designed to solve optimization problems (MaOPs) with numerous objectives. DSFMO prioritizes diversity over convergence, using a global diversity measure and a conditional convergence measure to ensure a balanced evolution process.
Similarly, \cite{weise2021scalable} presents DSFMO for general MaOPs, focusing on improving evolutionary algorithms, particularly for complex Pareto fronts. While that method uses a graph-based representation, our approach emphasizes the influence of the flow field on the agent's dynamics.


More specifically to the problem we address here, the literature offers a variety of approaches to tackling environmental flow dynamics. These methods include graph-based methods~\cite{Kularatne2018GoingWT} and evolutionary algorithms~\cite{Alvarez2004Evolutionary}, which prioritize energy-efficient path planning, as well as sampling-based planners~\cite{to2019streamlines}, which focus on achieving time-optimal paths. It is important to highlight that in scenarios with constant thrust (velocity of the vehicle concerning the flow), minimizing energy is equivalent to reducing time. In contrast, the travel time is directly proportional to the path length in constant net speed situations.

A comprehensive overview of trajectory planning and obstacle avoidance techniques for Autonomous Underwater Vehicles (AUVs) is presented in \cite{cheng2021path}. It focuses on algorithms that address the unique constraints and characteristics of AUVs, as well as the influence of marine environments. The article categorizes trajectory planning methods into two groups: global planning with known static obstacles, and local planning with unknown and dynamic obstacles.

The work \cite{yoo2021path} aims to develop a path-planning framework for marine robots that is robust to the inherent uncertainty in ocean current predictions. The focus is on using ensemble forecasts, which provide a distribution of possible flow fields, to generate path plans that minimize overall trajectory error. In contrast, our work presents a more general methodology for optimizing time and energy in static flow fields.

The paper \cite{kong20213d} describes an algorithm for estimating three-dimensional ocean flow fields using ensemble forecast data and online measurements. The proposed methodology leverages the property of negligible vertical velocity to produce highly accurate results, improving the trajectory planning capability of underwater gliders.
It uses singular value decomposition (SVD) and a Kalman filter for online updates, focusing on underwater applications. At the same time, we propose a more general approach to trajectory planning in static flow fields, employing a semi-Lagrangian method based on dynamic programming and genetic algorithms.

In \cite{yu2024learning}, the authors develop a reinforcement learning method for path planning in Autonomous Underwater Vehicles (AUVs) that maximizes data collection, such as temperature and salinity, while considering energy constraints and the influence of ocean currents. The focus is on balancing the acquisition of valuable information and minimizing energy consumption.

We propose a multi-objective approach that accounts for both the dynamic influence of environmental forces and the constraints imposed by fixed obstacles, making it suitable for applications where resource efficiency and obstacle avoidance are critical. We model the problem as an optimal control problem and introduce a novel technique called Harmonic Transformation, which is solved using a semi-Lagrangian scheme.

}

\section{Problem formulation}
\label{sec:problem}

Given a compact region of interest in the state space, we consider an agent described by:
\begin{align}
    \dot{\xv} &= \mathbf{f}(\xv,\uv), \label{eq:sys_dynamics}\\
    \dot{\xv} &= \mathbf{f}_1(\xv) + F_2(\xv)\uv, \label{eq:sys_affine}
\end{align}
\noindent with $\xv \in \reais^n$ being the agent's states, and $\uv \in \Uset$ the control inputs in the allowable control inputs set $\Uset \subset \reais^m$. Also, $\mathbf{f}(\xv,\uv) : \reais^{n+m} \rightarrow \reais^n$ represents the agent's dynamics, which can be decomposed into $\mathbf{f}_1(\xv): \reais^n \rightarrow \reais^n$ (the \emph{flow} vector field) and $\mathbf{F}_2(\xv) : \reais^n \rightarrow \reais^{n \times m}$ (the \emph{steering} matrix).

In the context of optimal control, we consider a cost function, $\ell(\xv,\uv) : \reais^{n+m} \rightarrow \reais$, that attributes a cost to every pair $(\xv,\uv)$ whose $\xv$ is not a \emph{target state}. We can also define a \emph{value function} $v(\xv): \reais^{n} \rightarrow \reais$ describing the minimum value for each point, $\xv$, in state space (with a trajectory starting at this point), such that:
\begin{align}
    v(\xv) = \inf_{\uv \in \Uset} \int_{0}^{\infty} \ell(\xv,\uv) dt. 
    \label{eq:value_original}
\end{align}
Since any path starting from a point along the optimal path should be optimal, we can use a Dynamic Programming Principle for $v(\xv)$ of the form:
\begin{align}
    v(\xv) = \inf_{\uv \in \Uset} \left\{ v(\yv_x(\Delta t, \uv)) + \int_0^{\Delta t}\ell(\yv_x(t, \uv),\uv) dt \right\},
    \label{eq:value_function}
\end{align}
with $\yv_x(t, \uv)$ representing the point at time $t$ along the path, taken when considering the control input defined by $\uv(t)$, for the system dynamics in \eqref{eq:sys_dynamics}.

To deal with constraints on $\xv$, such as forbidden or dangerous zones, we consider that the value function must be infinite in these locations. In addition, assuming that $\ell(\xv,\uv)$ is always non-negative, the value of the target location must always be null. When considered together, these constraints lead to the boundary conditions:
\begin{align}
    v(\xv) =
    \begin{cases}
        0 & \textrm{for } \xv(t) = \xv_g, \\ 
        \infty & \textrm{for } \xv(t) \in \partial \Obstacles, 
    \end{cases}
    \label{eq:boundary_conditions}
\end{align}
with $\xv_g$ representing a desired target location, and $\partial \Obstacles$ representing the boundaries of forbidden regions $\Obstacles$.

\subsection{Harmonic Transformation and Dynamic Programming} \label{subsect:Harmonic_Transform}

To properly deal with obstacles/forbidden zones, we transform the value function image from $[0,\infty)$ to $[0,1)$. Although the \emph{Kruzkov Transformation} is usually employed in these cases, it can lead to numerical problems when the value function assumes large values \cite{Falcone2013}. This problem can be somewhat mitigated by employing a scalar multiplying $v(\xv)$, but finding a suitable scalar can be a tiresome process in some cases.
In this sense, we consider a transformation of the form:
\begin{equation*}
    \bar{v}(\xv) = \mathcal{H}(v(\xv)) = \dfrac{v(\xv)}{1 + v(\xv)} = 1 - \dfrac{1}{1 + v(\xv)},
\end{equation*}
\noindent hereinafter referred to as \emph{Harmonic Transformation}, $\mathcal{H}(.)$.

As we shall detail subsequently, to derive a suitable Dynamic Programming Principle for the transformed problem, the following property of this transformation is essential:
\begin{align}
    \mathcal{H}(x_1 + x_2) &= \dfrac{x_1 + x_2}{1 + x_1 + x_2} \dfrac{\frac{1}{1 + x_1}}{\frac{1}{1 + x_1}}, \notag \\
    &= \dfrac{\mathcal{H}(x_1) + \frac{x_2}{1 + x_1} + x_2 - x_2}{\mathcal{H}(x_1) + \frac{1 + x_2}{1 + x_1}+ (1 + x_2) - (1 + x_2)}, \notag \\
    &= \dfrac{\mathcal{H}(x_1) + x_2 - x_2\left(1 - \frac{1}{1 + x_1}\right)}{1 + \mathcal{H}(x_1) + x_2 - (1 + x_2)\left(1 - \frac{1}{1 + x_1}\right)}, \notag \\
    &= \dfrac{\mathcal{H}(x_1) + \left(1 - \mathcal{H}(x_1)\right)x_2}{1 + \mathcal{H}(x_1) + x_2 - (1 + x_2)\mathcal{H}(x_1)}, \notag \\
    \mathcal{H}(x_1 + x_2) &= \dfrac{\mathcal{H}(x_1) + \left(1 - \mathcal{H}(x_1)\right)x_2}{1  + \left(1 - \mathcal{H}(x_1)\right)x_2}.
    \label{eq:harmonic_sum}
\end{align}

\subsubsection{Numerical approximation}

By applying property \eqref{eq:harmonic_sum}, a Dynamic Programming Principle can be found for the transformed value function:
\begin{align}
    \bar{v}(\xv(t)) &= \inf_{\uv \in \Uset} \left\{\dfrac{\bar{v}(\yv_x(\Delta t, \uv)) + q(\xv(t),\uv)}{1 + q(\xv(t),\uv)} \right\} \label{eq:DPP_harmonic}, \\
    q(\xv(t),\uv) &= \left(1 - \bar{v}(\yv_x(\Delta t, \uv))\right)\int_0^{\Delta t}\ell(\yv_x(t, \uv),\uv) dt, \notag
\end{align}
\noindent and a \ace{SL} numerical scheme can approximate the solution by employing a time discretization, followed by a space discretization. 

We consider the time discretization of Eq.~\eqref{eq:DPP_harmonic}, with step $\Delta t$, by applying a trapezoidal approximation for the integral term
\begin{align}
    \int_0^{\Delta t}\ell(\yv_x(t, \uv),\uv) dt &\approx \dfrac{\Delta t}{2} \left(\ell(\xv_{k+1},\uv_k) + \ell(\xv_k,\uv_k) \right), \notag \\
    g(\xv_k,\uv_k) &= \dfrac{\Delta t}{2} \left(\ell(\xv_{k+1},\uv_k) + \ell(\xv_k,\uv_k) \right), \label{eq:trap_int}
\end{align}
and a trapezoidal method to solve the system of equations composed of \eqref{eq:sys_dynamics} (assuming that the control input is held constant between the two samples), leading to
\begin{align}
    \xv_{k+1} &= \xv_k + \dfrac{\Delta t}{2} \left(\mathbf{f}(\xv_k,\uv_k) + \mathbf{f}(\xv_{k+1},\uv_{k})\right), \label{eq:trap_sys} \\
    \bar{v}_k(\xv_k)&= \inf_{\uv \in \Uset} \left\{\dfrac{\bar{v}_{k+1}(\xv_{k+1}) \!+\! \left(1 \!-\! \bar{v}_{k+1}(\xv_{k+1})\right)g(\xv_k,\uv_k)}{1 + \left(1 - \bar{v}_{k+1}(\xv_{k+1})\right)g(\xv_k,\uv_k)} \right\}. \label{eq:dpp__harmonic_discrete}
\end{align}

Next, we perform the space discretization of $\bar{v}$ by considering an unstructured grid of points covering the state space. Once these points represent $\bar{v}$, they are the only points over the space for which the value is updated. And since $\bar{v}(\xv_{k+1})$ might not be a part of the grid, it is replaced by a finite element \emph{linear} interpolation over the grid. 
One way of doing this linear interpolation is by employing a Delaunay triangulation on the unstructured grid points to find a triangulation of the space. We consider these triangles as our finite elements and represent the interpolation of $\bar{v}(\xv_{k+1})$ as $\mathcal{I}_{\overline{v}_{k+1}}[\xv_{k+1}]$. 

Taken together, both discretization (time and space) lead to an \ac{SL} approximation scheme of \eqref{eq:DPP_harmonic}, in the form:
\begin{equation}
    \hspace{-6pt}
    \resizebox{.92\linewidth}{!}{$
    \bar{v}_k(\xv_k) = \inf_{\uv \in \Uset} \left\{\dfrac{\mathcal{I}_{\bar{v}_{k+1}}[\xv_{k+1}] + \left(1 - \mathcal{I}_{\bar{v}_{k+1}}[\xv_{k+1}]\right)g(\xv_k,\uv_k)}{1 + \left(1 - \mathcal{I}_{\bar{v}_{k+1}}[\xv_{k+1}]\right)g(\xv_k,\uv_k)} \right\},$}
    \hspace{-4pt}
    \label{eq:dpp__harmonic_SL}
\end{equation}
with boundary conditions
\begin{equation}
    \bar{v}_k(\xv_{k}) = 
    \begin{cases}
        0 &\textrm{if } \xv(t) = \xv_g, \\ 
        1 &\textrm{if } \xv(t) \in \partial \Obstacles.
    \end{cases} 
    \label{eq:boundary}
\end{equation}

As in most approaches based on \ac{DP}, Eq.~\eqref{eq:dpp__harmonic_SL} can be solved backward in time using a \emph{value iteration} algorithm. Since we have formulated our problem as a stationary/infinite horizon optimal control problem, an acceleration technique known as \emph{policy iteration} \cite[Section 8.4.7]{Falcone2013} can also be employed. 

At every grid point, the optimal policy is:
\begin{align}
   \uv_k \!=\! \underset{\uv_k \in \Uset}{\operatorname{argmin}}\! \left\{\dfrac{\mathcal{I}_{\bar{v}}[\xv_{k+1}] + \left(1\!-\!\mathcal{I}_{\bar{v}}[\xv_{k+1}]\right)g(\xv_k,\uv_k)}{1 + \left(1 - \mathcal{I}_{\bar{v}}[\xv_{k+1}]\right)g(\xv_k,\uv_k)} \right\}, \label{eq:opt_policy_harmonic}
\end{align}
and fixed for this iteration. Afterward, the value function is updated according to conditions \eqref{eq:boundary} and
\begin{align}
    g(\xv_k,\uv_k) &= \bar{v}(\xv_{k}) \left(1 + \left(1 - \mathcal{I}_{\bar{v}}[\xv_{k+1}]\right)g(\xv_k,\uv_k)\right) \\& \quad - (1 - g(\xv_k,\uv_k))\mathcal{I}_{\bar{v}}[\xv_{k+1}]. 
    \label{eq:up_value_harmonic}
\end{align}
The algorithm iterates until it converges to the minimum of the value function.
Finally, the original value function \eqref{eq:value_original} can be recovered by:
\begin{equation*}
    v(\xv_{k}) = \dfrac{\bar{v}(\xv_{k})}{1 - \bar{v}(\xv_{k})}.
\end{equation*}

One of the main problems in this case is that Eq.~\eqref{eq:up_value_harmonic} defines a set of nonlinear equations. However, it is important to note that this is a highly sparse problem, for which specialized solvers drastically increase performance.

The \ac{DP} Principle in Eq.~\eqref{eq:DPP_harmonic} can be recast as a partial differential equation, known as the \ac{HJB} equation for the transformed value function of the form:
\begin{align}
    \sup_{\uv \in \Uset} \left\{-\nabla\bar{v}(\xv)\cdot\mathbf{f}(\xv,\uv) - \left(1 - \bar{v}(\xv)\right)^2 \ell(\xv,\uv) \right\} = 0.
    \label{eq:HJB_harmonic}
\end{align}

With this representation in mind, considering the usual regularity conditions in the literature to ensure that the value function is continuous \rev{\cite{Soner1986}}, we can state the following result regarding the optimal control problem and the proposed \ac{SL} approximation scheme.
%

\begin{thm} 
    \label{thm:convergence}
    Consider the optimal control problem represented by the \ac{HJB} equation \eqref{eq:HJB_harmonic}. As long as $\mathbf{f}$ and $\ell$ are Lipschitz, and $\ell$ is positive definite, with regards to the states, there exists a unique viscosity solution to this equation, representing the transformed value function of the optimal control problem. In addition to this, if $\ell > 0$, for every $\xv$ different from the target state, $\ell$ is convex with regards to $\uv$, $\uv$ is bounded, and $\mathbf{f}$ is bounded, the proposed numerical scheme converges to this unique solution as the time step, $\Delta t$, and the maximum distance between points on the grid, $\Delta x$, tend to zero, so long as $\Delta x$ tends faster than $\Delta t$.
\end{thm}
\begin{proof}
    The proof of this theorem is presented in \ref{app:Proof_thm}.
\end{proof}

\rev{
\begin{rem} \label{rem:convergence}
    Theorem \ref{thm:convergence} ensures that, as $\Delta x \rightarrow 0$ and $\Delta t \rightarrow 0$ (with $\Delta t > \Delta x$), the Semi-Lagrangian numerical solution converges to the unique viscosity solution of the corresponding \ac{HJB}. In addition to this, its proof shows that, so long as $\ell > 0$, for every $\xv$ different from the target state, the numerical scheme is \emph{monotone} and a \emph{contraction mapping} (even if the other assumptions of the theorem are not met, or the viscosity solution is discontinuous). This ensures that the method always converges to its unique fixed point, from the Banach Fixed Point Theorem, uniformly (since it is monotone). If \emph{value iteration} is employed directly, by solving equation \eqref{eq:dpp__harmonic_SL} backward in time, the distance between the current \emph{value function} to the optimal one can be shown to decay, in the best scenario, as
    \begin{align}
        \|\bar{v}_k - \bar{v}^{*}\| \leq \left(\dfrac{1}{1 + \bar{g}\Delta t}\right)^2 \|\bar{v}_{k+1} - \bar{v}^{*}\|, 
    \end{align}
    and in the worst scenario, as
    \begin{align}
        \|\bar{v}_k - \bar{v}^{*}\| \leq \dfrac{1}{1 + \varepsilon\bar{g}\Delta t} \|\bar{v}_{k+1} - \bar{v}^{*}\|,
    \end{align}
    with $1-\varepsilon$ the largest value, smaller than 1, that the optimal value assumes. In either case, we have a decay rate that inversely depends on the size of the time step ($\Delta t$). \emph{Policy iteration}, on the other hand, can usually be shown to converge \emph{superlinearly} (with a convergence rate between $1.5$ and $2$) with a decay rate that inversely depends on the maximum distance between points on the grid ($\Delta x$)\cite{Santos2004Convergence}.
\end{rem}
}

\subsection{Energy and time value functions} 

Even though we have shown that an \ac{SL} scheme can be employed with the \emph{Harmonic Transformation} with any cost function $\ell$ that satisfies the conditions of Thm.~\ref{thm:convergence}, in this work, we will focus our attention on two specific costs, and the multi-objective problem defined by them.

Assuming that
\begin{equation}
    \ell_T(\xv,\uv) = 
    \begin{cases}
        1 & \textrm{if } \xv \neq \xv_g, \\ 
        0 & \textrm{if } \xv = \xv_g,\\    
    \end{cases}
    \label{eq:cost_time}
\end{equation}
\noindent its value function in Eq.~\eqref{eq:value_original} can be written as:
\begin{equation}
    v_T(\xv) = \inf_{\uv \in \Uset} \int_{0}^{\infty} \!\ell_T(\xv,\uv) dt = \inf_{\uv \in \Uset} \int_{0}^{T(\xv)} \!1 dt = T(\xv),
    \label{eq:time_value}
\end{equation}
\noindent which corresponds to the \emph{time taken to reach the target state from the current state}, hereinafter referred to as \emph{time value function}.

We aim to examine the cost incurred in the following form:
\begin{equation*}
    \ell(\xv,\uv) = \left\{\begin{array}{ll} 
    \uv^T \uv, & \textrm{if } \xv \neq \xv_g, \\ 
    0, & \textrm{if } \xv = \xv_g,\\
    \end{array} \right.
\end{equation*}
\noindent so that its value function represents the \emph{energy required to reach the target state from the current state} (assuming the energy is given by the squared $\mathcal{L}_2$ norm of the control inputs), it could lead to an \emph{ill-posed} optimal control problem since the agent could decide to stop \emph{indefinitely} at any \emph{equilibrium point} of its dynamics, without being penalized.

In that regard, we consider the cost:
\begin{equation}
    \ell_E(\xv,\uv) = 
    \begin{cases}
        \varepsilon + \uv^T \uv & \textrm{if } \xv \neq \xv_g, \\ 
        0 & \textrm{if } \xv = \xv_g,\\
    \end{cases}
    \label{eq:cost_energy}
\end{equation}
with $\varepsilon$ being a small scalar that penalizes the time spent to reach the target state. Consequently, its value function in Eq.~\eqref{eq:value_original} can be written as:
\begin{align}
     v_E(\xv) &= \inf_{\uv \in \Uset} \int_{0}^{\infty} \!\!\!\!\ell_E(\xv,\uv) dt = \inf_{\uv \in \Uset} \int_{0}^{T(\xv)} \!\!(\varepsilon + \uv^T \uv) dt, \notag \\
     &= \varepsilon T(\xv) + \inf_{\uv \in \Uset} \int_{0}^{T(\xv)} \!\!\uv^T \uv\; dt, \label{eq:energy_value}
\end{align}
\noindent and can be interpreted as \emph{the energy used to reach the target state plus a small penalization of the time taken to reach it from the current state}. We will refer to this value function as the \emph{energy value function}.

Since these two cost functions (and respective value functions) are antagonistic, in this work we try to find a set of \emph{Pareto efficient solutions} (described either by the corresponding value function or policy defined for the agent), in a multi-objective manner. In that regard, we present two approaches to this multi-objective optimization: a \emph{deterministic} one (based on solving multiple single-objective optimizations concurrently), and an \emph{evolutionary} one (based on a proper multi-objective evolutionary algorithm).

\section{Proposed approaches}
\label{sec:proposed_approaches}

\subsection{Concurrent Policy Iteration}

A simple and direct way of dealing with this multi-objective optimal control problem would be to perform a scalarization of the cost function:
\begin{equation}
    \ell_\alpha(\xv,\uv) = \alpha \ell_T(\xv,\uv) + (1\!-\!\alpha) \ell_E(\xv,\uv), ~~ \alpha \in [0,1],
    \label{eq:cost_alpha}
\end{equation}
\noindent which leads to the suitable value function in Eq.~\eqref{eq:value_original}. A grid could be performed on this \emph{convex scalar weight} $\alpha$, which would lead to different mono-objective problems that could be solved separately and correspond to different efficient solutions on the Pareto set of optimal solutions for this problem.

One way to improve this proposal is by noting that, despite solving for different value functions, all these optimal control problems explore the same system (described by the agent's dynamics, target states, and obstacles/forbidden regions). In that regard, the policies could be evaluated under all of the single-objective costs/value functions simultaneously leading to a higher \emph{exploration} of the optimal control problem.

In this setting, a \emph{naive} implementation, considering policy iteration and a grid of $n_p$ scalar points in $\alpha$, would consist of updating a policy for each grid point, but calculating all $n_p$ value functions for each of these points. The main advantage of this approach is that it allows, in a sense, the different policies to share information employing the many value functions considered. The main drawback is that it requires the update of $n_p^2$ value functions at each time step, and this is the most costly step in many cases.

If we were not dealing with transformed value functions, it would suffice to consider the time and energy value functions in \eqref{eq:time_value} and \eqref{eq:energy_value}, respectively, since the rest of the value functions could be composed by taking the convex combination of them using $\alpha$. Since we are dealing with value functions transformed by the proposed \emph{Harmonic Transformation}, we need to define a relationship between them, and the transformation of a convex combination of these two value functions. In that regard, we can write
\begin{align}
    &\mathcal{H}(\alpha v_T + (1\!-\!\alpha)v_E) = \dfrac{\alpha v_T + (1\!-\!\alpha)v_E}{1+\alpha v_T + (1\!-\!\alpha)v_E} \notag \\
    &\quad = \dfrac{\alpha \frac{\mathcal{H}(v_T)}{1-\mathcal{H}(v_T)} + (1\!-\!\alpha)\frac{\mathcal{H}(v_E)}{1-\mathcal{H}(v_E)}}{1 + \alpha \frac{\mathcal{H}(v_T)}{1-\mathcal{H}(v_T)} + (1\!-\!\alpha)\frac{\mathcal{H}(v_E)}{1-\mathcal{H}(v_E)}} \notag \\
    &\quad = \dfrac{\alpha\mathcal{H}(v_T) + (1\!-\!\alpha)\mathcal{H}(v_E) - \mathcal{H}(v_T)\mathcal{H}(v_E)}{1 - (1\!-\!\alpha)\mathcal{H}(v_T) - \alpha\mathcal{H}(v_E)}. \label{eq:harmonic_convex_combination}
\end{align}

By applying Eq.~\eqref{eq:harmonic_convex_combination}, we can find any of the transformed value functions (for a given policy) as long as we have the transformed time and energy value functions. So, only two value functions are updated and stored for every policy, and only $2n_p$ value functions are updated at each step, leading to the \acd{CPI} procedure presented in Algorithm~\ref{alg:cpi}.

\rev{
\begin{rem}
    \emph{Policy Iteration} schemes ensure that, at every iteration, a better policy is found. Therefore, if a discretization of the set of allowable controls is considered (such that the minimum can be found from an inspection over the set), they are assured to converge in a finite number of steps (since there will only be a finite number of allowable policies). In the worst case, the number of iterations taken to converge can be exponential, with $\mathcal{O}\left(\dfrac{n_u^{n_x}}{n_x}\right)$, $n_u$ the number of points in our allowable controls set, and $n_x$ the number of points discretizing our state space \cite{Hollanders2012Exponential}. As discussed in Remark \ref{rem:convergence}, though, these methods exhibit \emph{superlinear} convergence and usually converge with a small number of iterations \cite{Santos2004Convergence}. In comparison to the single objective policy iteration, \acd{CPI} require $2 n_p$ times more \emph{policy evaluations} (lines 4, 5, 11 and 12 in Algorithm \ref{alg:cpi}) and $n_p$ times more \emph{policy improvement} (line 10 in Algorithm \ref{alg:cpi}) steps.
\end{rem}
}

\begin{algorithm}[!ht]
    \caption{\protect\ace{CPI}}
    \label{alg:cpi}
    \begin{algorithmic}[1]    
        \REQUIRE $\xv_g$, $\bm{\alpha}$, $\varepsilon$
        \STATE $k \gets 0$
        \FORALL{$\alpha_i \in \bm{\alpha}$}
            \STATE Assign a corresponding initial policy $\uv_0^{(i)}$
            \STATE Calculate the transformed time value $\bar{v}_{T0}^{(i)}$ of using the $\uv_0^{(i)}$ policy, the cost in \eqref{eq:cost_time}, and \eqref{eq:trap_sys}, \eqref{eq:trap_int} and \eqref{eq:up_value_harmonic}
            \STATE Calculate the transformed energy value $\bar{v}_{E0}^{(i)}$ of using the $\uv_0^{(i)}$ policy, the cost in \eqref{eq:cost_energy}, and \eqref{eq:trap_sys}, \eqref{eq:trap_int} and \eqref{eq:up_value_harmonic}
        \ENDFOR
        \WHILE{policies $\uv_k^{(i)}$ have not converged}
            \STATE $k \gets k+1$
            \FORALL{$\alpha_i \in \bm{\alpha}$}
                \STATE Find the \emph{best} current value for $\alpha_i$ by {\small 
                \begin{align*}
                    &\tilde{v}^{(i)}(\xv) = \\
                    &\quad \min_s \dfrac{\alpha_i \bar{v}_{T(k-1)}^{(s)} + (1\!-\!\alpha_i)\bar{v}_{E(k-1)}^{(s)} - \bar{v}_{T(k-1)}^{(s)}\bar{v}_{E(k-1)}^{(s)}}{1 - (1\!-\!\alpha_i)\bar{v}_{T(k-1)}^{(s)} - \alpha_i\bar{v}_{E(k-1)}^{(s)}}
                \end{align*}}
                \STATE Updates the policy by using the cost \eqref{eq:cost_alpha} defined by $\alpha_i$ using \eqref{eq:trap_sys}, \eqref{eq:trap_int} and {\small $$
                \uv_k^{(i)} = \underset{\uv_k \in \Uset}{\operatorname{argmin}} \left\{\dfrac{\mathcal{I}_{\tilde{v}}[\xv_{k+1}] + \left(1 - \mathcal{I}_{\tilde{v}}[\xv_{k+1}]\right)g(\xv_k,\uv_k)}{1 + \left(1 - \mathcal{I}_{\tilde{v}}[\xv_{k+1}]\right)g(\xv_k,\uv_k)} \right\}
                $$}
                \STATE Calculate the transformed time value $\bar{v}_{Tk}^{(i)}$ using the $\uv_k^{(i)}$ policy, cost in \eqref{eq:cost_time}, and \eqref{eq:trap_sys}, \eqref{eq:trap_int} and \eqref{eq:up_value_harmonic}
                \STATE Calculate the transformed energy value $\bar{v}_{Ek}^{(i)}$ using the $\uv_k^{(i)}$ policy, cost in \eqref{eq:cost_energy}, and \eqref{eq:trap_sys}, \eqref{eq:trap_int} and \eqref{eq:up_value_harmonic}
            \ENDFOR
        \ENDWHILE
        \RETURN Policies $\uv_k^{(i)}$, and value functions $\bar{v}_{Tk}^{(i)}$ and $\bar{v}_{Ek}^{(i)}$
    \end{algorithmic}
\end{algorithm}

\subsection{Genetic Algorithm and Evolutionary Policy Iteration}

While the proposed \rev{\ace{CPI}} algorithm exchanges information about the single-objective problems being solved to obtain better and faster results, it still consists of solving $n_p$ single-objective optimal control problems. In addition to this, to find the optimal policies, it is common to consider a finite set of allowable control inputs (so that the minimum can be found from a simple verification of all possible elements).

To overcome both problems, here we propose a multi-objective genetic algorithm. To deal with the problem in a multi-objective approach directly, we employ the NSGA-II~\cite{Deb2002} (specifically its \emph{fast non-dominated sorting} and \emph{crowding distance} procedures), whereas the Evolutionary Policy Iteration \cite{Chang2005,Chang2013} approach is adopted to deal with a large/continuous set of allowable controls.

In this paper, we use policies $\uv$ to represent the individuals in the population. Equation \eqref{eq:up_value_harmonic} is employed with costs \eqref{eq:cost_time} and \eqref{eq:cost_energy} to associate time and energy value functions to these policies, respectively. Note that the optimal policy, when considering any value function, would lead to a smaller value over the grid of unstructured points (since it needs to be the best policy everywhere inside of our region of interest). In that regard, we use the average of the value function over the unstructured grid as a proxy for how good that policy is, and end up with a bi-objective optimization problem with average time value and average energy value as our costs. Since we are employing an \emph{Harmonic Transformation} on both value functions, they assume values in $[0,1]$ and, as such, so will their average values.

In a single-objective setting, \emph{Evolutionary Policy Iteration} \cite{Chang2005} makes use of the concept of \emph{Policy Switching} to generate an elite individual which is guaranteed to be an improvement over the policies in the current population. In our setting, \emph{Policy Switching} can be written, with regards to a specific value function, as
\begin{align}
    \tilde{v}(\xv_k) \! &= \!\!\min_{i \in \textrm{population}} \!\!v_i(\xv_k), \notag\\
    \uv_{ps}(\xv_k) &=\!\! \underset{\uv_k \in \Uset_{pop}(\xv_k)}{\operatorname{argmin}} \!\! \left\{\!\dfrac{\mathcal{I}_{\tilde{v}}[\xv_{k+1}] \!+\! \left(1 \!-\! \mathcal{I}_{\tilde{v}}[\xv_{k+1}]\right)g(\xv_k,\uv_k)}{1 \!+\! \left(1 \!-\! \mathcal{I}_{\tilde{v}}[\xv_{k+1}]\right)g(\xv_k,\uv_k)} \!\right\},\notag
\end{align}
\noindent with $\tilde{v}(\xv_k)$ representing the value function obtained by taking the minimum value over the individuals of the population, $\Uset_{pop}(\xv_k)$ the set of controls defined by the policies in the population at point $\xv_k$, and $\uv_{ps}(\xv_k)$ the policy defined by taking the best control available in $\Uset_{pop}(\xv_k)$ according to $\tilde{v}(\xv_k)$. \emph{Policy Switching} can also be employed as a \emph{cross-over} operator that can generate an offspring given a set of parents, in which case a subset of the population is usually employed as the parents. In a single-objective setting, it can be shown that when paired with a \emph{mutation} operator to ensure the exploration of the policy space, \emph{policy switching} ensures the elite policy convergence to the optimal policy with probability one, when the state space is finite \cite{Chang2013}.

In our multi-objective setting, we can still employ \emph{policy switching} to generate elite individuals. Since every individual already possesses time and energy value functions associated with its policy, it is straightforward to generate elite individuals concerning time and energy. In addition to this, considering a scalar parameter $\alpha \in [0,1]$, we can employ \eqref{eq:harmonic_convex_combination} to recover elite individuals which are not on the extreme points of the Pareto set. 
In that regard, in our proposed algorithm, we generate three elite individuals using \emph{policy switching} at every generation, the two extreme policies (associated with minimum time and minimum energy), and a third one using a random $\alpha$ at every generation. Even though the value of $\alpha$ could be fixed, at $\alpha = 0.5$ for instance, we employ a random $\alpha$ to increase exploration of the Pareto Set. To increase the convergence of the method, as well as ensure some diversity in the population, this elite individual with a random $\alpha$ also considers some control possibilities from a fixed set of control actions (aside from the ones on the individuals on the population) when searching for the value of the best action.

Similarly, \emph{policy switching} is also employed as a \emph{cross-over} operator, with a random $\alpha \in [0,1]$, to generate other new offspring in the population. In this case, however, only a small number of individuals (usually 2 or 3) are employed as parents. To create new individuals, part of the offspring is also created by employing a simple \emph{cross-over} between two policies. A random scalar value $\lambda$ \rev{is sampled from a uniform distribution over $[0,1]$} and given policies $\uv_1(\xv_k)$ and $\uv_2(\xv_k)$, a new policy is generated by
\begin{align}
    \uv_{\textrm{off}} = \lambda \uv_1(\xv_k) + (1 - \lambda) \uv_2(\xv_k). 
    \label{eq:simple_xover}
\end{align}
With both \emph{cross-over} operators, a simple \emph{mutation} with a Gaussian noise (but modified to respect the bounds of the allowable control set) is employed in these new individuals to ensure exploration of the policy space.

As in a standard NSGA-II algorithm, the \emph{fast non-dominated sorting} procedure is employed to rank solutions according to dominance in the objective functions space, whereas the \emph{crowding distance} is employed to rank solutions on the same level (which do not dominate one another). These ranking solutions are employed when selecting individuals from the population (both for choosing parents for the \emph{cross-over} operators and for selecting survivors for the next generation). Parents are chosen according to a binary tournament, whereas survivors (from the combined original and offspring population) are chosen according to their level of dominance. For the last positions available in the surviving population, a roulette strategy is employed, in which the probability of selection is proportional to the \emph{crowding-distance} of an individual.
This approach is summarized in Algorithm~\ref{alg:MEPI}.

\begin{algorithm}
    \caption{\small\protect\ac{MEPI}}
    \label{alg:MEPI}
    \begin{algorithmic}[1]    
        \REQUIRE $\xv_g$, $n_{\textrm{pop}}$, $n_{\textrm{ger}}$, $n_{\textrm{cp}}$, $n_{\textrm{par}}$, $\varepsilon$
        \STATE $k \gets 0$
        \FORALL{individual $i$ in the population}
            \STATE Assign a random initial policy $\uv^{(i)}$
            \STATE Calculate the transformed time value $\bar{v}_{T}^{(i)}$ and transformed energy value $\bar{v}_{E}^{(i)}$ using the $\uv^{(i)}$ policy, as well as their average values over the state space
        \ENDFOR
        \WHILE{$k < n_{\textrm{ger}}$}
            \STATE $k \gets k+1$
            \STATE Calculate \emph{crowding distance} of the current population
            \STATE Use \emph{Policy Switching} over all  population to generate elite individuals with regards to the $\bar{v}_{T}^{(i)}$ and $\bar{v}_{E}^{(i)}$ value functions respectively
            \STATE Considering a random $\alpha$, use \emph{Policy Switching} over all the population to generate an elite individual with regards to the value functions
            \begin{displaymath}
                \tilde{v}^{(i)}(\xv) = \dfrac{\alpha_i \bar{v}_{T}^{(i)} + (1\!-\!\alpha_i)\bar{v}_{E}^{(i)} - \bar{v}_{T}^{(i)}\bar{v}_{E}^{(i)}}{1 - (1\!-\!\alpha_i)\bar{v}_{T}^{(i)} - \alpha_i\bar{v}_{E}^{(i)}}
            \end{displaymath}
            considering an \emph{additional} set of fixed control actions.
            \STATE Generate $n_{\textrm{cp}}$ new offspring individuals using \emph{Policy Switching} with $n_{\textrm{par}}$ parents chosen using a binary tournament (decided by \emph{dominance} and \emph{crowding distance}) and a Gaussian mutation. 
            \STATE Generate $n_{\textrm{pop}} - n_{\textrm{cp}} - 3$ individuals using the simple \emph{cross-over} in \eqref{eq:simple_xover} with 2 parents chosen using a binary tournament (decided by \emph{dominance} and \emph{crowding distance}) and a Gaussian mutation. 
            \STATE Considering the individuals from the original population and the offspring population, sort them using the \emph{fast non-dominated sorting} and calculate their \emph{crowding distance}
            \STATE Select individuals to form the next generation of the population according to their \emph{dominance level}. Employ a roulette (using the \emph{crowding distance}) to fill the available spots when the remaining individuals cannot be chosen with dominance only.
        \ENDWHILE
        \RETURN Population $\uv^{(i)}$, and value functions $\bar{v}_{T}^{(i)}$ and $\bar{v}_{E}^{(i)}$
    \end{algorithmic}
\end{algorithm}

\rev{
\begin{rem}
    Similarly to other stochastic optimal path planning methods in the literature, such as RRT$^\star$ \cite{Karaman2011}, \emph{Evolutionary Policy Iteration} can be shown to be asymptotically optimal (converge to the optimal value with probability 1) \cite{Chang2013}. In the same fashion, \acd{MEPI} can also be shown to be asymptotically optimal, since over time \emph{Policy Switching} will ensure better policies are found for the elite individuals (lines 8 and 9 in Algorithm \ref{alg:MEPI}). In addition to this, the use of the \emph{crowding distance} and \emph{fast non-dominated sorting} will ensure that the points found on the Pareto set of efficient solutions are well spread out. Unlike, \ac{CPI}, the additional cost for running \ac{MEPI} over the single-objective equivalent is not that high (only doubling the required \emph{Policy Evaluation} steps). Note that, since we only have asymptotical optimality, it can be hard to define a stopping criterion for \ac{MEPI}, and the number of generations is usually employed.
\end{rem}
}

\section{Numerical simulations}
\label{sec:numSim}
\rev{To illustrate the methods proposed in this paper, this section presents 5 numerical examples. All examples were run on a Ryzen 7 2700 CPU with 16GB of RAM on Windows 11 with MATLAB R2023a. To solve \eqref{eq:up_value_harmonic}, we employed the \emph{fsolve} function with the \emph{`SpecifyObjectiveGradient'} option and made use of sparse matrices to describe the Jacobian matrix. 

Throughout this section, whenever we refer to the \emph{Average Time Cost} and \emph{Average Energy Cost} as a way to compare the value functions, we will employ an average over the points of the CPI/MEPI grid of the transformed time, $\bar{v}_T$, and transformed energy, $\bar{v}_E$. These are interesting metrics, as using the Harmonic Transformation guarantees that they always assume values in $[0,1]$. Note that, the use of the Harmonic Transformation implies that these are dimensionless quantities, and as such do not possess a corresponding unit.}
\subsection{\rev{Example 1 - comparing Harmonic and Kruzkov transformations}}
\begin{figure}
    \centering
    \includegraphics[width=\linewidth]{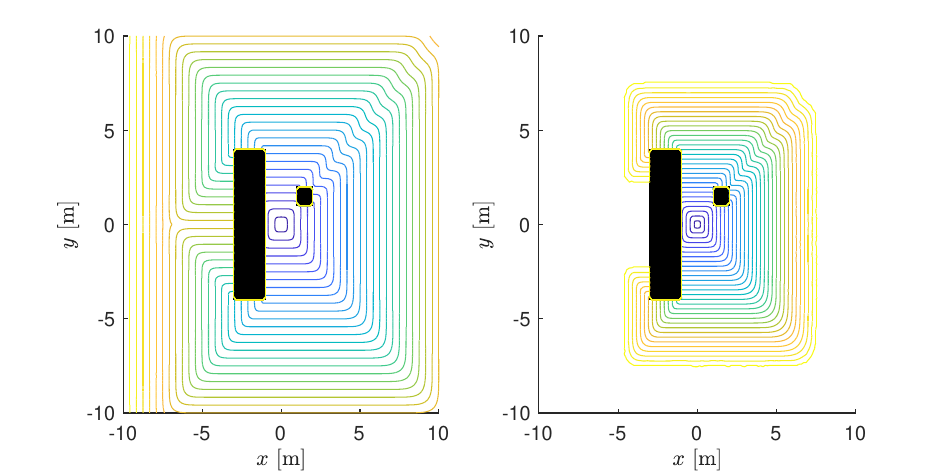}
    \caption{\rev{Time to reach the origin employing the \emph{Harmonic} transformation and the \emph{Kruzkov} transformation in Example 1. The obstacles are presented in \textcolor{black}{black} whereas the level curves of the \emph{time to reach the origin} value functions are presented as the colored curves. The left plot represents the solution found by the \emph{Harmonic} transformation while the right plot represents the solution found by the \emph{Kruzkov} transformation.}}
    \label{fig:ex0Contorno}
\end{figure}

\begin{figure}
    \centering
    \includegraphics[width=\linewidth]{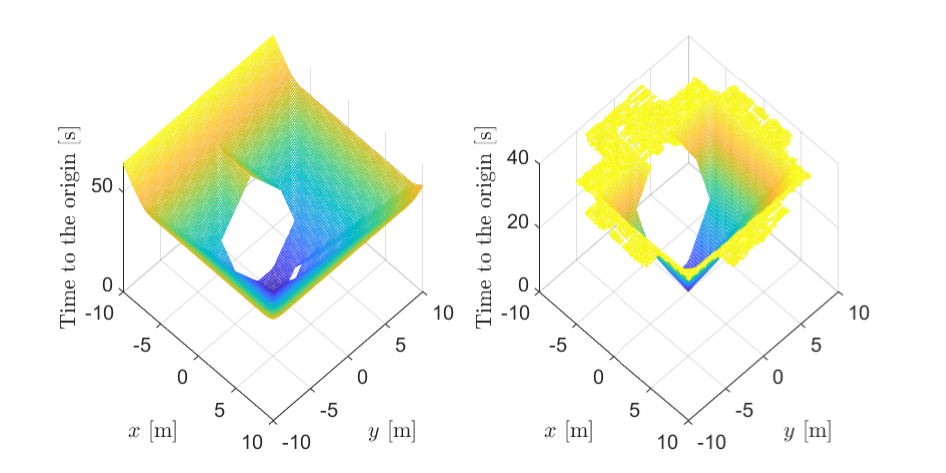}
    \caption{\rev{Time to reach the origin employing the \emph{Harmonic} transformation and the \emph{Kruzkov} transformation in Example 1. The left plot represents the solution found by the \emph{Harmonic} transformation while the right plot represents the solution found by the \emph{Kruzkov} transformation.}}
    \label{fig:ex0Valor}
\end{figure}

\rev{In this first example, we consider a simple scenario to compare the numerical behavior of the Harmonic transformation against the Kruzkov transformation. We consider an agent without drift dynamics (\emph{i.e.} without a flow field affecting its velocities), whose behavior can be described by
\begin{align}
    \begin{bmatrix} \dot{x} \\ \dot{y} \end{bmatrix} = \begin{bmatrix} v_x \\ v_y \end{bmatrix},
\end{align}
\noindent with $x$ and $y \in [-10, 10]$ representing the position on the plane, and $v_x$ and $v_y \in [-0.2, 0.2]$ the agent's velocities, used as control inputs. The control objective is to drive the agent from any state $(x,y)$ to the origin while avoiding the obstacles. 

Since the main purpose of this example is to compare the numerical properties of both transformations, in this single example, only the minimum time problem is considered, and both implementations employ only a value iteration scheme (\emph{i.e.} the Harmonic transformation scheme is implemented by solving \eqref{eq:dpp__harmonic_SL}, whereas the Kruzkov transformation scheme is implemented by solving \cite[Eq. (8.69)]{Falcone2013}) with no acceleration schemes in either case. A structured grid with 141 points in $X$ and 141 points in $y$, leading to 19881 points, was employed, with a time step of $\Delta t = 1s$. 

To illustrate the numerical behavior of each transformation, the value functions found by each method are converted back to the \emph{time to reach the origin} value function. This can be done by $v_T = \dfrac{\bar{v}}{1 - \bar{v}}$ for the Harmonics transformation, and $v_T = -\ln(1-\bar{v})$ for the Kruzkov transformation. These are presented in Figures \ref{fig:ex0Contorno} and \ref{fig:ex0Valor}. As can be seen in these figures, the Harmonic transformation was capable of representing the value function throughout the whole domain, whereas the Kruzkov transformation gets numerically unstable once the time to reach the origin exceeds $40s$ (being unable to cover the whole domain). This illustrates the problem discussed in Section~\ref{subsect:Harmonic_Transform}, justifying the proposition of the Harmonic transformation.
}

\subsection{Example \rev{2 - simple linear drift dynamics}}

\rev{In this second example}, we consider an agent with simple linear drift dynamics
\begin{align}
    \begin{bmatrix} \dot{x}_1 \\ \dot{x}_2 \\ \dot{x}_3 \end{bmatrix} &= 
    \begin{bmatrix} -1 & 1.2094 & 0.6937 \\ -1.2094 & -1 & 2.6564 \\ -0.6937 & -2.6564 & -1 \end{bmatrix} 
    \begin{bmatrix} x_1 \\ x_2 \\ x_3 \end{bmatrix} \notag  \\&+ 
    \begin{bmatrix} -0.2415 & 0.3971 & 0.8855 \\ -0.9701 & -0.0744 & -0.2312 \\ -0.0259 & -0.9148 & 0.4031 \end{bmatrix}
    \begin{bmatrix} v_1 \\ v_2 \\ v_3 \end{bmatrix},
\end{align}
\noindent with $x_1$, $x_2$ and $x_3 \in [-1, 1]$ being the position on the space, and $v_1$, $v_2$ and $v_3 \in [-2, 2]$ the control inputs. The control objective is to drive the agent from any state $(x_1,x_2,x_3)$ to the goal position $(-0.2, 0.2, 0)$. \rev{Note that, from the system dynamics, the flow vector field is described as 
\begin{align}
    \mathbf{f}(\mathbf{x}) = \begin{bmatrix} -1 & 1.2094 & 0.6937 \\ -1.2094 & -1 & 2.6564 \\ -0.6937 & -2.6564 & -1 \end{bmatrix} 
    \begin{bmatrix} x_1 \\ x_2 \\ x_3 \end{bmatrix}
\end{align}
and represent the \emph{drift} dynamics, whereas the term
\begin{equation}
    \begin{bmatrix} -0.2415 & 0.3971 & 0.8855 \\ -0.9701 & -0.0744 & -0.2312 \\ -0.0259 & -0.9148 & 0.4031 \end{bmatrix}
    \begin{bmatrix} v_1 \\ v_2 \\ v_3 \end{bmatrix}
\end{equation}
represent the \emph{steering} terms for this linear system.} 

This problem was solved using the \ac{CPI} (Alg.~\ref{alg:cpi}) and the \ac{MEPI} (Alg.~\ref{alg:MEPI}), and in both cases, the space was discretized by a structured grid of 1332 points (with the grid points equally distributed in space, and with the goal added to the grid). The step size used for the Trapezoidal method was $\Delta t = 0.1$.

\begin{figure}
    \centering
    \includegraphics[width=\linewidth]{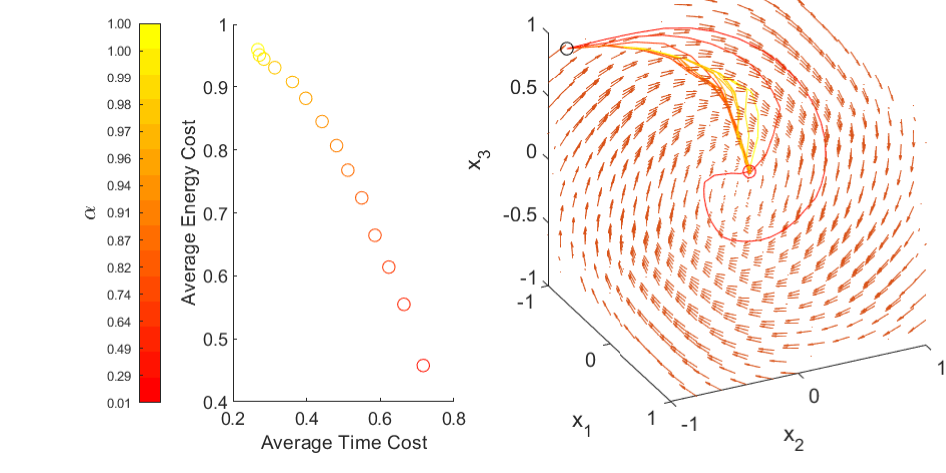}
    \caption{Result running the \emph{Concurrent Policy Iteration} for Example \rev{2}. The left plot represents the solution set found, while the right plot shows how each solution corresponds to a different path in state space, for initial point $(-0.9, -0.9, 0.9)$. The \textcolor{red}{red arrows} represent the flow vector field, and the solutions are color-coded to match each trajectory in state space with a point on the objective function space, with \textcolor{yellow}{yellow} being the fastest trajectory and \textcolor{red}{red} the trajectory that spends the least amount of energy.}
    \label{fig:ex1CPI}
\end{figure}

For the \ac{CPI}, we considered 14 logarithmically spaced values of $\alpha_i$ ranging from $0.01$ to $1.0$, and the allowable inputs were discretized considering 9 points in each direction ($v_1$, $v_2$ and $v_3$), leading to 729 possible control actions. The algorithm ran 15 iterations to find the result presented in Fig.~\ref{fig:ex1CPI}.

\begin{figure}
    \centering
    \includegraphics[width=\linewidth]{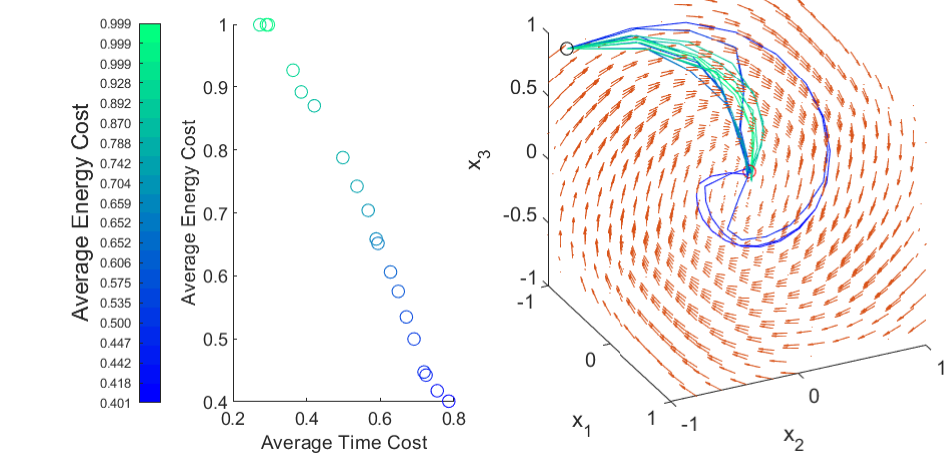}
    \caption{Result running the \emph{Multi-objective Evolutionary Policy Iteration} for Example \rev{2}. The left plot represents the final population found while the right plot shows how each solution corresponds to a different path in state space, for initial point $(-0.9, -0.9, 0.9)$. The \textcolor{red}{red arrows} represent the flow vector field, and the solutions are color-coded to match each trajectory in state space with a point on the objective function space, with \textcolor{green}{green} being the fastest trajectory and \textcolor{blue}{blue} the trajectory that spends the least amount of energy.}
    \label{fig:ex1MEPI}
\end{figure}

For the \ac{MEPI}, we considered a population of 20 individuals, with $n_{cp} = 15, n_{par} = 3$ and $\sigma = 0.2$ as the standard deviation for the Gaussian mutation. The initial population was randomly chosen using a uniform distribution over the allowable controls. The algorithm ran for 60 generations to find the result presented in Figure \ref{fig:ex1MEPI}.

For this example, both algorithms were able to find similar solutions. It is important to note that, even though for this example the \ac{CPI} was able to find a set of solutions that seem well-distributed in the objective space, that is not always the case. Finding a suitable set of $\alpha_i$ values can take some time. This problem does not happen with \ac{MEPI} since it ranks solutions based on the \emph{crowding distance} metric.

\subsection{Example \rev{3} - scenario with obstacles}

As a \rev{third} example, we consider an agent whose drift dynamics approximate a vortex with constant flow velocity centered at $(0.5,0)$, and complete actuation. The dynamics can be described by:
\begin{align}
    \bar{\mathbf{f}}(x,y) &= 
    \begin{bmatrix} 
        -1 & 3 \\ 
        -3 & -1 
    \end{bmatrix} 
    \begin{bmatrix} 
        x-0.5 \\ 
        y 
    \end{bmatrix}, \notag\\
    \begin{bmatrix} 
        \dot{x} \\ 
        \dot{y} 
    \end{bmatrix} &= \dfrac{1}{0.01 + \|\bar{\mathbf{f}}(x,y)\|_2} \bar{\mathbf{f}}(x,y) + 
    \begin{bmatrix} 
        v_x \\ 
        v_y 
    \end{bmatrix},
\end{align}
\noindent with $x$ and $y \in [-1, 1]$ representing the position on the plane, and $v_x$ and $v_y \in [-2, 2]$ the agent's velocities \rev{relative to} the drift field, used as control inputs. The control objective is to drive the agent from any state $(x,y)$ to the goal position $(-0.5, 0.6)$ while avoiding the obstacle. 
Once again this problem was solved by employing both, \ac{CPI} and \ac{MEPI}. The space was originally discretized by 596 points, however, with this discretization, the \ac{CPI} was not capable of finding suitable solutions. Therefore, in this example, the \ac{MEPI} used 596 points to discretize the state space, while \ac{CPI} used 796 points (in both cases these points include the desired goal position and 95 points describing the boundaries of the obstacle). The step size of the trapezoidal method used was $\Delta t = 0.05$ for both methods.

\begin{figure}
    \centering
    \includegraphics[width=\linewidth]{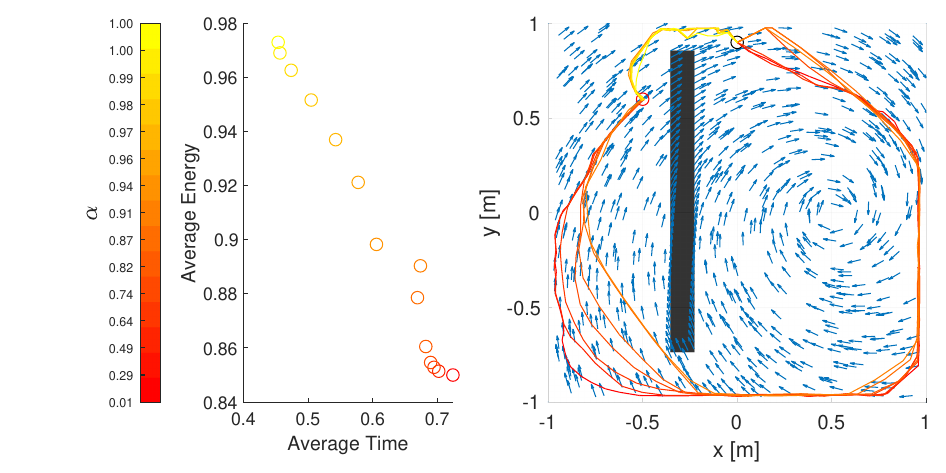}
    \caption{Result found when running \emph{Concurrent Policy Iteration} for Example \rev{3}. The left plot represents the solutions found while the right plot shows how each solution corresponds to a different path in state space, for initial point $(0, 0.9)$. The \textcolor{blue}{blue arrows} represent the flow vector field, the \textbf{black region} represent an obstacle, and the solutions are color-coded to match each trajectory in state space with a point on the objective function space, with \textcolor{yellow}{yellow} being the fastest trajectory and \textcolor{red}{red} the trajectory that spends the least amount of energy.}
    \label{fig:ex2CPI}
\end{figure}

For the \ac{CPI}, we considered 15 logarithmically spaced values of $\alpha_i$ ranging from $0.01$ to $1$, and the allowable inputs were discretized considering 15 points in each direction ($v_x$ and $v_y$), leading to 225 possible control actions. The algorithm ran for 30 iterations to find the result presented in Figure \ref{fig:ex2CPI}.

\begin{figure}
    \centering
    \includegraphics[width=\linewidth]{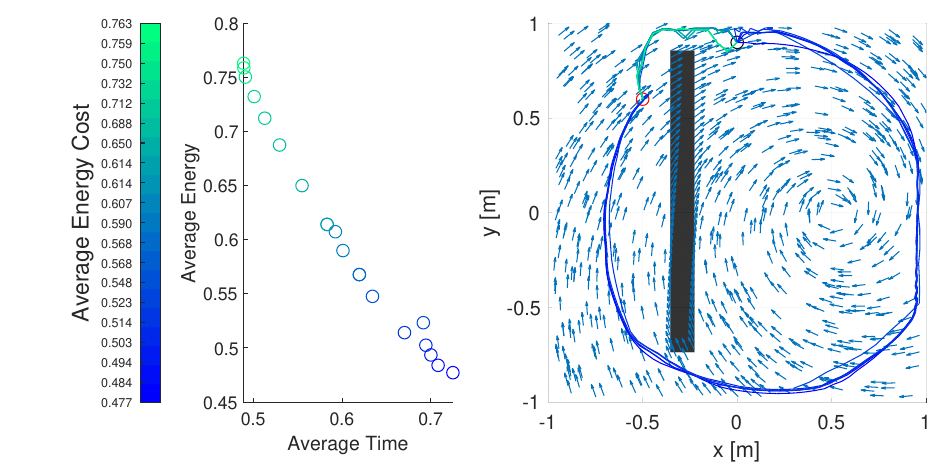}
    \caption{Result found when running \emph{Multi-objective Evolutionary Policy Iteration} for Example \rev{3}. The left plot represents the final population found while the right plot shows how each solution corresponds to a different path in state space, for initial point $(0, 0.9)$. The \textcolor{blue}{blue arrows} represent the flow vector field, the \textbf{black region} represent an obstacle, and the solutions are color-coded to match each trajectory in state space with a point on the objective function space, with \textcolor{green}{green} being the fastest trajectory and \textcolor{blue}{blue} the trajectory that spends the least amount of energy.}
    \label{fig:ex2MEPI}
\end{figure}

For \ac{MEPI}, we considered a population with 20 individuals, with $n_{cp} = 15, n_{par} = 3$, and $\sigma = 0.2$ as the standard deviation for the Gaussian mutation. The initial population was randomly chosen using a uniform distribution over the allowable controls. The algorithm ran for 100 generations to find the result presented in Figure \ref{fig:ex2MEPI}.

For this example, unlike with Example 1, the solutions were considerably different (looking at the objective function space) and the solutions found with MEPI have a substantially smaller energy cost than the ones from CPI. This can be explained by the fact that CPI uses a discrete set of control actions, while MEPI can use a continuous set. It is also interesting to note that, with both methods, the optimal time paths chose the shortest path to the goal, while the energy optimal path tries to make the most use of the drift dynamics possible.

\rev{\subsubsection{Comparison with RRT$^\star$}}

\rev{Additionally, we have also employed another optimal path planning algorithm from the literature, RRT$^\star$ \cite{Karaman2011,webb2012kinodynamic,Zhang2022}. In this example, to generate paths from every point on the state space to the desired target (similarly to what our algorithms do), we employed a single tree starting from the desired target and evolving backwards in time. We ran the algorithm 15 times, with the same $\alpha_i$ values as in the CPI, corresponding to different scalar cost functions for each case. We considered an Euler time-discretization and the time step was reduced to $\Delta t = 0.025$ to compensate (concerning our use of a trapezoidal method in our methods). With this time-discretization employed, the neighborhood sets were calculated by the reachable sets over a single time step (forward reachable sets for the rewiring step, and backward reachable sets for choosing optimal parents) so that the RRT$^\star$ steering could be done via quadratic programming. To achieve similar accuracy as the methods presented in this paper, the RRT$^\star$ ran until the tree had 6000 points.}

\begin{figure}
    \centering
    \includegraphics[width=\linewidth]{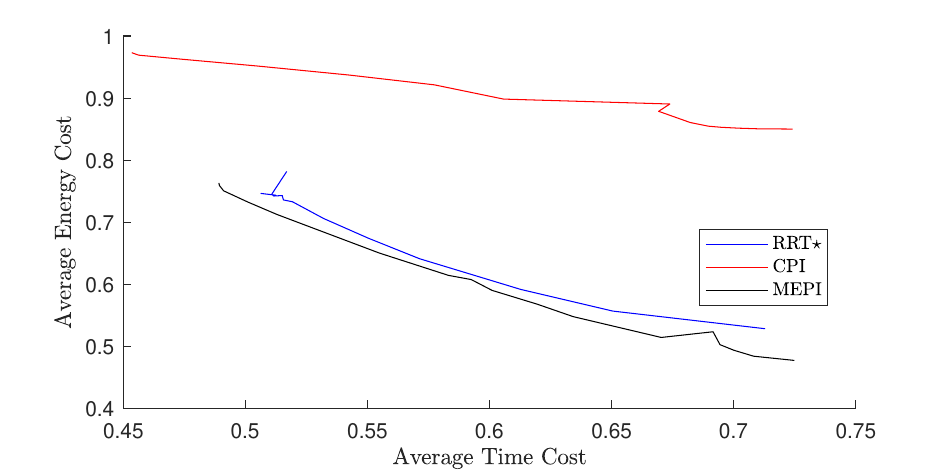}
    \caption{\rev{Comparison of the Pareto set found in Example 3 by employing an RRT$^\star$, \emph{Concurrent Policy Iteration} and \emph{Multi-objective Evolutionary Policy Iteration}. The \textcolor{blue}{blue line} indicates the RRT$^\star$ solutions, the \textcolor{red}{red line} indicates the CPI solutions and the \textcolor{black}{black line} indicates the MEPI solutions.}}
    \label{fig:ex2RRT}
\end{figure}

\rev{A comparison among the Pareto Sets found in this Example is presented in Figure \ref{fig:ex2RRT}. It can be seen that, for this particular example, the Pareto Set found using MEPI and RRT$^\star$ are quite similar, with the MEPI dominating the RRT$^\star$ solutions by a small margin. The CPI solutions are almost completely dominated (aside from the solutions near the minimum time solution) by the other approaches. As previously discussed, this can be explained by the fact that CPI uses a discrete set of control actions whereas the other two approaches employ a continuous set of control actions, which seems to be a key aspect in finding minimum energy solutions for this particular example.}

\subsection{Example \rev{4} - marine navigation problem}

In this \rev{fourth} example, we consider a problem in which the agent can be regarded as a marine vessel moving through an ocean environment. In this case, the flow vector field describes the ocean currents, whereas the steering matrix describes the allowable input velocity directions for the vehicle.

\begin{figure}
    \centering
    \includegraphics[width=\linewidth]{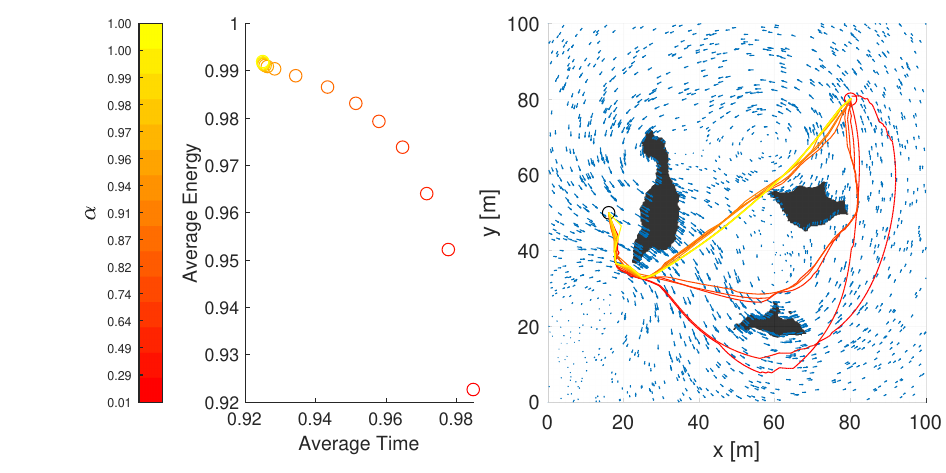}
    \caption{Result found when running \emph{Concurrent Policy Iteration} for Example \rev{4}. The left plot represents the solutions found while the right plot shows how each solution corresponds to a different path in state space, for the initial point $(16, 50)$. The \textcolor{blue}{blue arrows} represent the flow vector field, the \textbf{black regions} represent obstacles, and the solutions are color-coded to match each trajectory in state space (on the right) with a point on the objective function space (on the left), with \textcolor{yellow}{yellow} being the fastest trajectory and \textcolor{red}{red} the trajectory that spends the least amount of energy.}
    \label{fig:ex3CPI}
\end{figure}

\begin{figure}
    \centering
    \includegraphics[width=\linewidth]{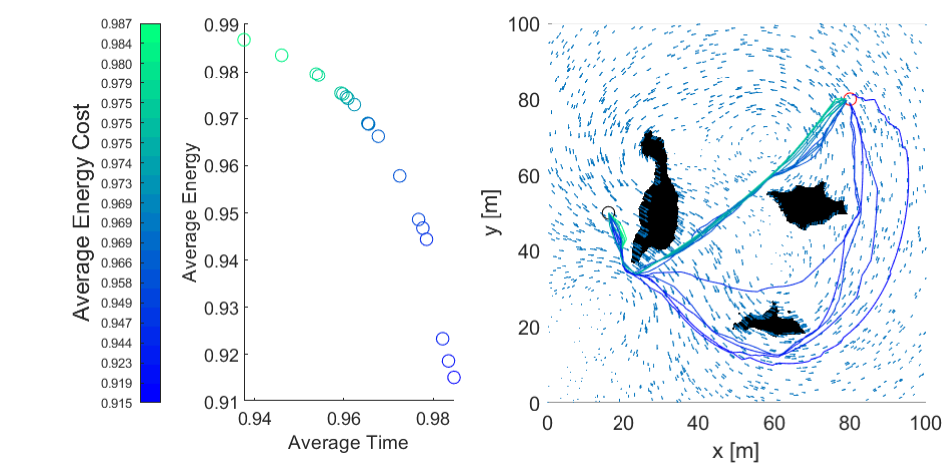}
    \caption{Result found when running \emph{Multi-objective Evolutionary Policy Iteration} for Example \rev{4}. The left plot represents the final population found while the right plot shows how each solution corresponds to a different path in state space, for the initial point $(16, 50)$. The \textcolor{blue}{blue arrows} represent the flow vector field, the \textbf{black regions} represent the obstacles, and the solutions are color-coded to match each trajectory in state space with a point on the objective function space, with \textcolor{green}{green} being the fastest trajectory and \textcolor{blue}{blue} the trajectory that spends the least amount of energy.}
    \label{fig:ex3MEPI}
\end{figure}

Inspired by \cite{Garau2006AUV,Mansfield2022Energy} we consider an ocean current model in $\reais^2$ given by the superposition of different one-point vortex solutions called \emph{viscous Lamb vortices}, given by
\begin{equation}
    \footnotesize
    \mathbf{f}_i(\xv) = \Gamma_i
    \begin{bmatrix}
        -\dfrac{y - c_{2i}}{2\pi(\xv - \mathbf{c}_i)^T (\xv - \mathbf{c}_i)} \left(1 - e^{-\frac{(\xv - \mathbf{c}_i)^T (\xv - \mathbf{c}_i)}{\delta_i^2}} \right)\\
         \dfrac{x - c_{1i}}{2\pi(\xv - \mathbf{c}_i)^T (\xv - \mathbf{c}_i)} \left(1 - e^{-\frac{(\xv - \mathbf{c}_i)^T (\xv - \mathbf{c}_i)}{\delta_i^2}} \right)
    \end{bmatrix},
\end{equation}
in which, $\xv = (x,y)$ is the vessel position, $\mathbf{c}_i = (c_{1i},c_{2i})$ is the center of the $i$-th vortex, and $\delta_i$ and $\Gamma_i$ are parameters related to its radius and strength. In this example, we consider that four vortices are used, so that the agent dynamics can be described by
\begin{align}
    \begin{bmatrix} \dot{x} \\ \dot{y} \end{bmatrix} = \left(\sum_{i=1}^4  \mathbf{f}_i(\xv) \right) + \begin{bmatrix} v_x \\ v_y \end{bmatrix},
\end{align}
with $v_x \in [-3, 3]$ and $v_y \in [-3, 3]$ being the input velocities, and parameters $\Gamma_1 = -50$, $\Gamma_2 = \Gamma_3 = \Gamma_4 =  50$, $\delta_1 = \delta_2 = \delta_3 = \delta_4 = 10$, $\mathbf{c}_1 = (20,30)$, $\mathbf{c}_2 = (60,70)$, $\mathbf{c}_3 = (27,65)$, and $\mathbf{c}_4 = (60,30)$. The environment is a 100m x 100m square ($x \in [0, 100]$, $y \in [0, 100]$) with island shaped objects and is depicted in Figures \ref{fig:ex3CPI} and \ref{fig:ex3MEPI}. The goal is to drive the vessel from any point $(x,y)$ in this area to the goal position $(80,80)$ while avoiding the obstacles.

Similarly to the other examples, this problem was solved by employing CPI and MEPI. In both cases, the space was discretized by an unstructured grid of 3412 points and the step size used for the Trapezoidal method was $\Delta t = 1$. 

For CPI, we considered 15 logarithmically spaced values of $\alpha_i$ ranging from $0.01$ to $1$, and the allowable inputs were discretized considering 15 points in each direction ($v_x$ and $v_y$), leading to 225 possible control actions. The algorithm ran for 30 iterations to find the result presented in Figure \ref{fig:ex3CPI}.

For MEPI, we considered a population with 20 individuals, with $n_{cp} = 15, n_{par} = 3$ and $\sigma = 0.2$ as the standard deviation for the Gaussian mutation. The initial population was randomly chosen using a uniform distribution over the allowable controls. The algorithm ran for 60 generations to find the result presented in Figure \ref{fig:ex3MEPI}.

Similarly to Example 1, the solutions found in this example were comparable for both methods (though the concentration of points in the objective space was a little different in both cases). When considering the extremes of the Pareto Set of solutions found, CPI was able to find the best time-optimal solution, whereas MEPI was able to find the best energy-optimal solution.

It is also interesting to note that, with both methods, when moving along the efficient solutions they can be \emph{visually} divided into 3 groups of paths (considering the trajectories taken in a closed loop).

\subsection{Example \rev{5} - time-varying periodic flow field}

\begin{figure}
    \hspace{-2.5cm} \includegraphics[width=1.3\linewidth]{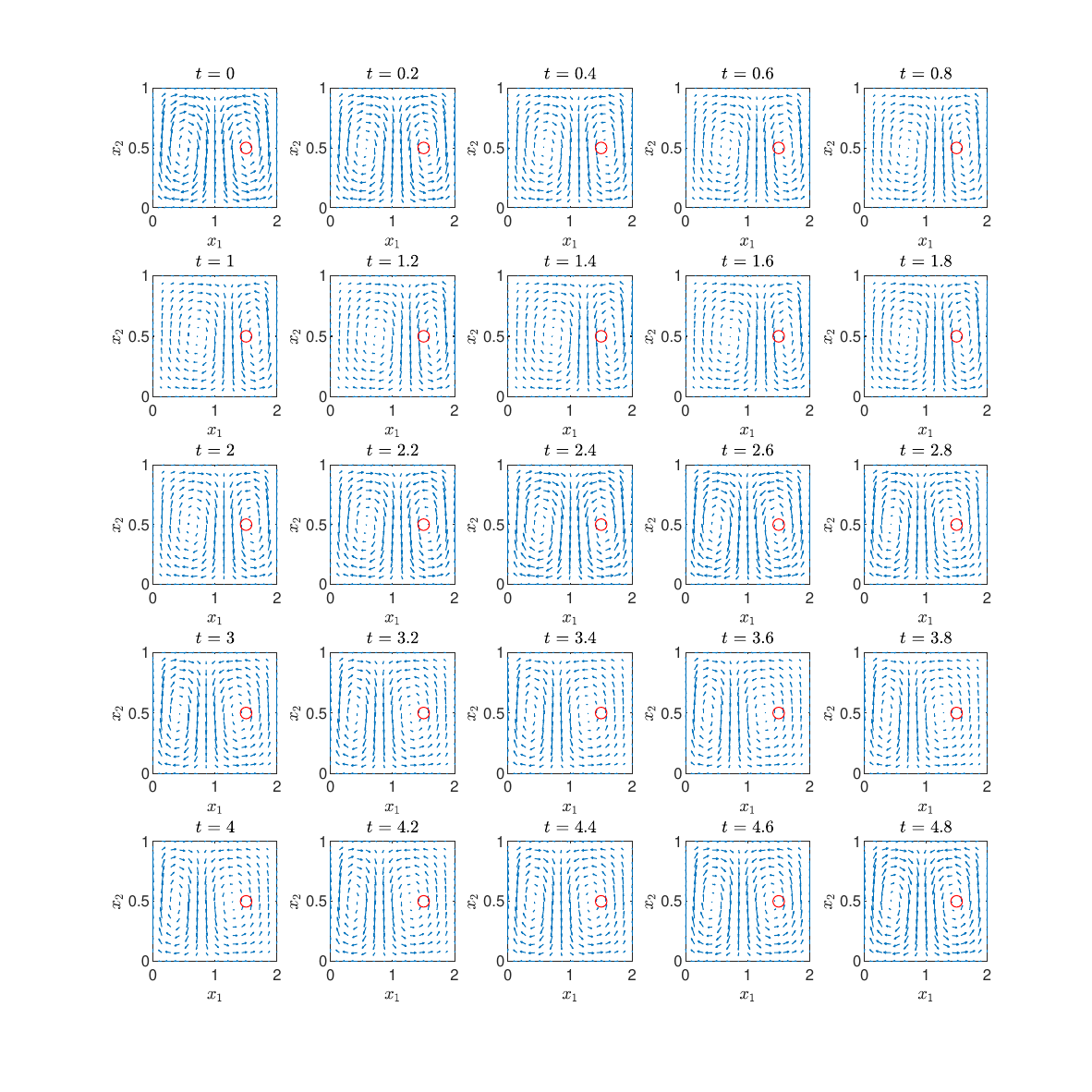}
    \caption{Time-varying flow field for \eqref{eq:ex4FF} in Example \rev{5}. Each plot represents a time snapshot of the flow field, and the \textcolor{red}{red} circle represents the desired goal in this example. Note that, even though the flow field is time-varying, it is periodic with a period of $5$, therefore the snapshots represent one period. Video: \url{https://youtu.be/ycnHhr4KtrQ}}
    \label{fig:ex4OL}
\end{figure}

Finally, In this last example, we consider a problem with a time-varying flow field, to illustrate that, even though the proposed approach was developed for static flow fields, it can still be employed in this case through a simple transformation of the state space. In this case, we consider that the flow field is a time-varying periodic double-gyre, given by
\begin{equation}
    \begin{bmatrix} \dot{x}_1 \\ \dot{x}_2 \end{bmatrix} = \begin{bmatrix} -\theta \pi \sin\left(\pi (a(t) x_1^2 + b(t) x_1)\right) \cos(\pi x_2) \\ \theta \pi (2 a(t) x_1 + b(t)) \cos\left(\pi (a(t) x_1^2 + b(t) x_1)\right) \sin(\pi x_2)  \end{bmatrix} + \begin{bmatrix} v_{x_1} \\ v_{x_2} \end{bmatrix},
    \label{eq:ex4FF}
\end{equation}
with $x_1 \in [0, 2]$ and $x_2 \in [0, 1]$ the state space coordinates, $v_{x_1} \in [-0.8, 0.8]$ and $v_{x_2} \in [-0.8, 0.8]$ being the input velocities, $a(t)$ and $b(t)$ periodic signals given by
\begin{align}
    a(t) &= \varepsilon \sin(\omega t), \\
    b(t) &= 1 - 2\varepsilon \sin(\omega t), \\
\end{align}
and parameters $\theta = 0.1$, $\varepsilon = 0.25$ and $\omega = \frac{2\pi}{5}$. The control objective in this example is to drive the agent from any state $(x_1,x_2)$ to the goal position $(1.5, 0.5)$. The time-varying flow field and the goal position are illustrated in Fig.~\ref{fig:ex4OL}.

Even though the methods presented in this paper were not directly developed to deal with time-varying flow fields, note that they can still deal with these problems if we augment the system's state space description to include the time $t$ as a state (therefore guaranteeing that all of the system's dynamics can be directly defined simply by the states and the control inputs. In that regard, for our implementation, we consider a dynamics defined as
\begin{equation}
    \begin{bmatrix} \dot{x}_1 \\ \dot{x}_2 \\ \dot{t} \end{bmatrix} = \begin{bmatrix} -\theta \pi \sin\left(\pi (a(t) x_1^2 + b(t) x_1)\right) \cos(\pi x_2) \\ \theta \pi (2 a(t) x_1 + b(t)) \cos\left(\pi (a(t) x_1^2 + b(t) x_1)\right) \sin(\pi x_2)  \\ 1 \end{bmatrix} + \begin{bmatrix} v_{x_1} \\ v_{x_2} \\ 0 \end{bmatrix}.
\end{equation}

Like the other examples before it, this problem was solved using the \ac{CPI} (Alg.~\ref{alg:cpi}) and the \ac{MEPI} (Alg.~\ref{alg:MEPI}), and in both cases, the space was discretized by a structured grid of 5625 points (with a $15 \times 15 \times 25$ grid over $x_1$, $x_2$ and $t$). The step size used for the Trapezoidal method was $\Delta t = 0.2$ (matching the step size on the grid for $t$). Since the flow field employed is periodic, time loops on itself every 5 seconds, and as such we only represent time over $[0,5)$.

\begin{figure}
    \centering
    \includegraphics[width=\linewidth]{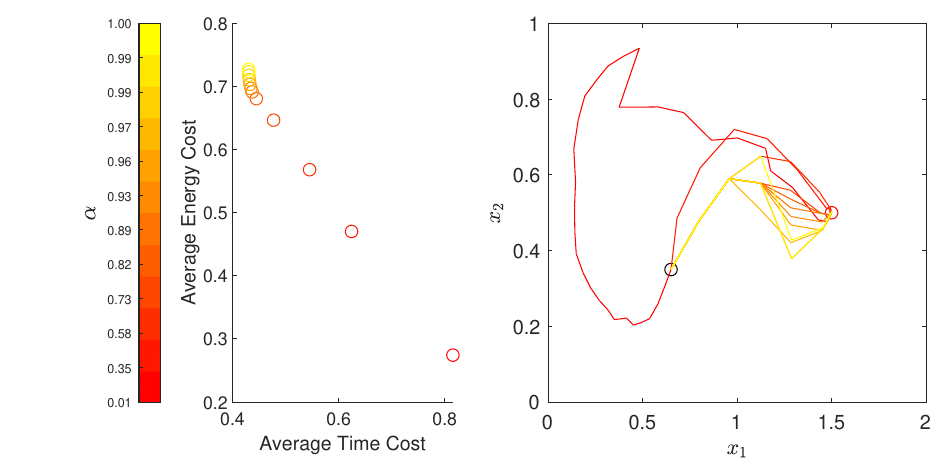}
    \caption{Result found when running \emph{Concurrent Policy Iteration} for Example \rev{5}. The left plot represents the solutions found while the right plot shows how each solution corresponds to a different path in state space, for initial point $(0.65, 0.35)$. The solutions are color-coded to match each trajectory in state space (on the right) with a point on the objective function space (on the left), with \textcolor{yellow}{yellow} being the fastest trajectory and \textcolor{red}{red} the trajectory that spends the least amount of energy.}
    \label{fig:ex4CPI}
\end{figure}

For CPI, we considered 12 logarithmically spaced values of $\alpha_i$ ranging from $0.01$ to $1$, and the allowable inputs were discretized considering 15 points in each direction ($v_{x_1}$ and $v_{x_2}$), leading to 225 possible control actions. The algorithm ran 14 iterations to find the result presented in Figure \ref{fig:ex4CPI}.

\begin{figure}
    \centering
    \includegraphics[width=\linewidth]{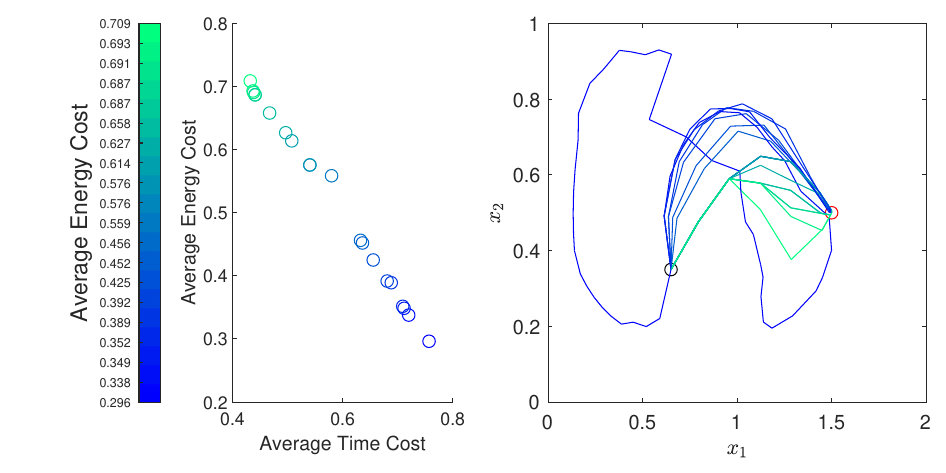}
    \caption{Result found when running \emph{Multi-objective Evolutionary Policy Iteration} for Example \rev{5}. The left plot represents the final population found while the right plot shows how each solution corresponds to a different path in state space, for the initial point $(0.65, 0.35)$. The solutions are color-coded to match each trajectory in state space with a point on the objective function space, with \textcolor{green}{green} being the fastest trajectory and \textcolor{blue}{blue} the trajectory that spends the least amount of energy.}
    \label{fig:ex4MEPI}
\end{figure}

For MEPI, we considered a population with 20 individuals, with $n_{cp} = 20, n_{par} = 3$ and $\sigma = 0.2$ as the standard deviation for the Gaussian mutation. The initial population was randomly chosen using a uniform distribution over the allowable controls. The algorithm ran for 30 generations to find the result presented in Figure \ref{fig:ex4MEPI}.

\begin{figure}
    \centering
    \includegraphics[width=\linewidth]{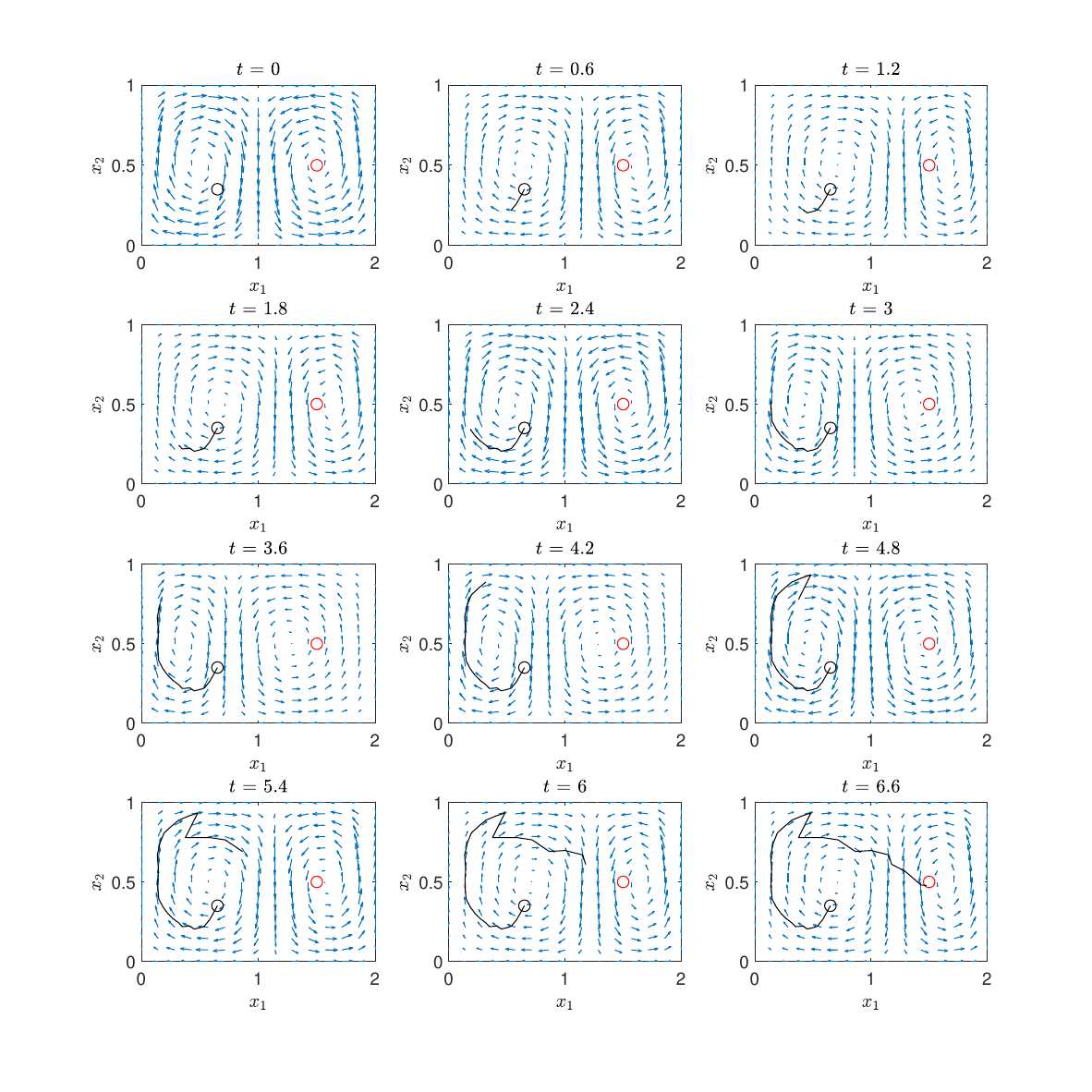}
    \caption{Time \emph{snapshots} of one of the trajectory solutions found by \emph{Concurrent Policy Iteration} in Example \rev{5}. Each plot represents a time snapshot of the time-varying flow field, in \textcolor{blue}{blue}, the trajectory is displayed in black, and the \textcolor{red}{red} circle represents the desired goal in this example. Video: \url{https://youtu.be/pOeKqx_r4u0}}
    \label{fig:ex4traj}
\end{figure}

Aside from the minimal energy solution (which looks a bit different for both methods), the trajectories found for both methods are quite similar. Unlike the previous examples, though, it is hard to see those trajectories over the flow field in a single plot. In that regard, Figure \ref{fig:ex4traj} illustrates one of the trajectories found over several time snapshots of the flow field.

\section{Conclusion and Future Work}
\label{sec:conc}

Path planning approaches are of paramount importance in Robotics, as they enable autonomous robots to navigate complex and dynamic environments with efficiency and safety.
This paper proposes a multi-objective path planning formulation to determine paths that simultaneously optimize travel time and energy consumption.

The presented methodology incorporates the dynamic influence of environmental flow fields and considers obstacles and forbidden zones. 
Our approach relies on the proposed Harmonic Transformation, which maps values onto a specific range, effectively mitigating potential numerical issues.

Here we presented two distinct approaches to determine the set of Pareto efficient solutions within the context of multi-objective optimization. The first approach is deterministic, involving the simultaneous solution of multiple single-objective optimizations. This deterministic method capitalizes on the parallel resolution of individual objectives to achieve Pareto efficiency. The second approach takes on an evolutionary perspective. By employing principles of evolution and selection, this evolutionary approach explores the solution space to uncover more adaptive and comprehensive Pareto optimal solutions.

\rev{As demonstrated in the examples, both methods were capable of solving the multi-objective optimal planning problem, though with different characteristics. \ac{CPI} usually demonstrated a faster convergence, in practice, and was better at finding solutions close to the time-optimal solution. However, since the optimization (in line 10 of Algorithm \ref{alg:cpi}) is done from a discretization of the set of allowable controls, its results were worse than \ac{MEPI} when considering the energy cost (especially in Example 2). \ac{MEPI}, on the other hand, was better suited to finding solutions close to the energy-optimal solution. Since it is only asymptotically optimal (meaning that it converges to the optimal value with probability 1), determining a suitable stop criterion is harder (a premature stopping of the method usually leads to suboptimal solutions).}

\rev{Even though the proposed approaches are capable of numerically solving the examples presented in this paper, they still suffer from the \emph{curse of dimensionality} since an increase in the number of dimensions usually demands an exponential increase in the number of points discretizing the state-space. As such, trying to overcome some of the computational limitations of the proposed approach, in future work, we intend to study: the use of sparse \cite{Kang2017} and adaptive grids \cite{munos2002variable,Grune2004}, to be capable of discretizing the state space with a reduced number of grid points; domain decomposition methods \cite{Festa2018}, to allow for faster parallel implementations as well as higher dimensional spaces; the use of \emph{Physics-Informed Neural Networks} as an alternative to the solution of our transformed HJB equations \cite{Meng2024,Shilova2024}, avoiding the need for the space discretization step; and Reinforcement Learning approaches \cite{Mukherjee2023}, in an actor-critic setting, which can be very interesting to deal with a continuous set of control actions.}


\section*{Acknowledgments}
This work has been financed by the \ac{CAPES} through the Academic Excellence Program (PROEX), \ac{CNPq} - grant numbers 310446/2021-0, 306286/2020-3, 407063/2021-8 and 308597/2023-0, and \ac{FAPEMIG} - grant number APQ-02228-22.

 \bibliographystyle{elsarticle-num} 
 \bibliography{references}

\begin{thebibliography}{10}
\expandafter\ifx\csname url\endcsname\relax
  \def\url#1{\texttt{#1}}\fi
\expandafter\ifx\csname urlprefix\endcsname\relax\def\urlprefix{URL }\fi
\expandafter\ifx\csname href\endcsname\relax
  \def\href#1#2{#2} \def\path#1{#1}\fi

\bibitem{LaValle2006Planning}
S.~M. LaValle, {Planning Algorithms}, Cambridge University Press, New York, NY, USA, 2006.

\bibitem{Hohmann2021Hybrid}
N.~Hohmann, M.~Bujny, J.~Adamy, M.~Olhofer, {Hybrid Evolutionary Approach to Multi-objective Path Planning for UAVs}, in: IEEE Symp. Series on Computational Intelligence (SSCI), 2021, pp. 1--8.

\bibitem{Cole2018Reactive}
K.~Cole, A.~M. Wickenheiser, Reactive trajectory generation for multiple vehicles in unknown environments with wind disturbances, IEEE Trans. on Robotics 34~(5) (2018) 1333--1348.

\bibitem{Weekly2014Autonomous}
K.~Weekly, A.~Tinka, L.~Anderson, A.~M. Bayen, Autonomous river navigation using the hamilton–jacobi framework for underactuated vehicles, IEEE Trans. on Robotics 30~(5) (2014) 1250--1255.

\bibitem{Falcone2013}
M.~Falcone, R.~Ferretti, Semi-Lagrangian approximation schemes for linear and Hamilton—Jacobi equations, SIAM, 2013.

\bibitem{Zafar2018Methodology}
M.~N. Zafar, J.~Mohanta, {Methodology for Path Planning and Optimization of Mobile Robots: A Review}, Procedia Computer Science 133 (2018) 141--152, international Conference on Robotics and Smart Manufacturing (RoSMa2018).

\bibitem{Low2019Solving}
E.~S. Low, P.~Ong, K.~C. Cheah, {Solving the optimal path planning of a mobile robot using improved Q-learning}, Robotics and Autonomous Systems 115 (2019) 143--161.

\bibitem{Gasparetto2015Path}
A.~Gasparetto, P.~Boscariol, A.~Lanzutti, R.~Vidoni, Path Planning and Trajectory Planning Algorithms: A General Overview, Springer International Publishing, Cham, 2015, pp. 3--27.

\bibitem{Foehn2021Timeoptimal}
P.~Foehn, A.~Romero, D.~Scaramuzza, Time-optimal planning for quadrotor waypoint flight, Science Robotics 6~(56) (2021) eabh1221.

\bibitem{Ahmed2013Multiobjective}
F.~Ahmed, K.~Deb, Multi-objective optimal path planning using elitist non-dominated sorting genetic algorithms, Soft Computing 17~(7) (2013) 1283--1299.

\bibitem{Xue2018Solving}
Y.~Xue, J.-Q. Sun, {Solving the Path Planning Problem in Mobile Robotics with the Multi-Objective Evolutionary Algorithm}, Applied Sciences 8~(9) (2018).

\bibitem{MA2018Multiobjective}
Y.~Ma, M.~Hu, X.~Yan, Multi-objective path planning for unmanned surface vehicle with currents effects, ISA Transactions 75 (2018) 137--156.

\bibitem{Macharet2021Minimal}
D.~G. Macharet, A.~{Alves Neto}, D.~Shishika, {Minimal Exposure Dubins Orienteering Problem}, IEEE Robotics and Automation Letters 6~(2) (2021) 2280--2287.

\bibitem{Mansfield2022Energy}
A.~Mansfield, D.~G. Macharet, M.~A. Hsieh, Energy-efficient orienteering problem in the presence of ocean currents, in: IEEE/RSJ Int. Conf. on Intelligent Robots and Systems (IROS), 2022, pp. 10081--10087.

\bibitem{Mansfield2023Energy}
A.~Mansfield, D.~G. Macharet, M.~A. Hsieh, Energy-efficient team orienteering problem in the presence of time-varying ocean currents, in: IEEE/RSJ Int. Conf. on Intelligent Robots and Systems (IROS), 2023.

\bibitem{Doshi2023EnergyTime}
M.~M. Doshi, M.~S. Bhabra, P.~F. Lermusiaux, {Energy–time optimal path planning in dynamic flows: Theory and schemes}, Computer Methods in Applied Mechanics and Engineering 405 (2023) 115865.

\bibitem{zhang2024many}
W.~Zhang, J.~Liu, Y.~Liu, J.~Liu, S.~Tan, A many-objective evolutionary algorithm under diversity-first selection based framework, Expert Systems with Applications 250 (2024) 123949.

\bibitem{weise2021scalable}
J.~Weise, S.~Mostaghim, A scalable many-objective pathfinding benchmark suite, IEEE Transactions on Evolutionary Computation 26~(1) (2021) 188--194.

\bibitem{Kularatne2018GoingWT}
D.~Kularatne, S.~Bhattacharya, M.~A. Hsieh, Going with the flow: a graph based approach to optimal path planning in general flows, Autonomous Robots 42 (2018) 1369--1387.

\bibitem{Alvarez2004Evolutionary}
A.~Alvarez, A.~Caiti, R.~Onken, Evolutionary path planning for autonomous underwater vehicles in a variable ocean, IEEE J. of Oceanic Engineering 29~(2) (2004) 418--429.

\bibitem{to2019streamlines}
K.~C. To, K.~M.~B. Lee, C.~Yoo, S.~Anstee, R.~Fitch, Streamlines for motion planning in underwater currents, in: Int. Conf. on Robotics and Automation (ICRA), IEEE, 2019, pp. 4619--4625.

\bibitem{cheng2021path}
C.~Cheng, Q.~Sha, B.~He, G.~Li, Path planning and obstacle avoidance for auv: A review, Ocean Engineering 235 (2021) 109355.

\bibitem{yoo2021path}
C.~Yoo, J.~J.~H. Lee, S.~Anstee, R.~Fitch, Path planning in uncertain ocean currents using ensemble forecasts, in: IEEE International Conference on Robotics and Automation (ICRA), IEEE, 2021, pp. 8323--8329.

\bibitem{kong20213d}
F.~H. Kong, K.~C. To, G.~Brassington, S.~Anstee, R.~Fitch, 3d ensemble-based online oceanic flow field estimation for underwater glider path planning, in: IEEE/RSJ International Conference on Intelligent Robots and Systems (IROS), IEEE, 2021, pp. 4358--4365.

\bibitem{yu2024learning}
Y.~Yu, H.~Zheng, W.~Xu, Learning and sampling-based informative path planning for auvs in ocean current fields, IEEE Transactions on Systems, Man, and Cybernetics: Systems (2024).

\bibitem{Soner1986}
H.~M. Soner, Optimal control with state-space constraint {I}, SIAM Journal on Control and Optimization 24~(3) (1986) 552--561.
\newblock \href {https://doi.org/10.1137/0324032} {\path{doi:10.1137/0324032}}.

\bibitem{Santos2004Convergence}
M.~S. Santos, J.~Rust, Convergence properties of policy iteration, SIAM J. on Control and Optimization 42~(6) (2004) 2094--2115.

\bibitem{Hollanders2012Exponential}
R.~Hollanders, J.-C. Delvenne, R.~M. Jungers, The complexity of policy iteration is exponential for discounted markov decision processes, in: IEEE Conf. on Decision and Control (CDC), 2012, pp. 5997--6002.

\bibitem{Deb2002}
K.~Deb, A.~Pratap, S.~Agarwal, T.~Meyarivan, A fast and elitist multiobjective genetic algorithm: {NSGA-II}, IEEE Trans. on Evolutionary Computation 6~(2) (2002) 182--197.

\bibitem{Chang2005}
H.~S. Chang, H.-G. Lee, M.~Fu, S.~Marcus, Evolutionary policy iteration for solving markov decision processes, IEEE Trans. on Automatic Control 50~(11) (2005) 1804--1808.

\bibitem{Chang2013}
H.~Chang, J.~Hu, M.~Fu, S.~Marcus, Simulation-Based Algorithms for Markov Decision Processes, Communications and Control Engineering, Springer London, 2013.

\bibitem{Karaman2011}
S.~Karaman, E.~Frazzoli, Sampling-based algorithms for optimal motion planning, The International Journal of Robotics Research 30~(7) (2011) 846--894.
\newblock \href {https://doi.org/10.1177/0278364911406761} {\path{doi:10.1177/0278364911406761}}.

\bibitem{webb2012kinodynamic}
D.~J. Webb, J.~van~den Berg, Kinodynamic {RRT*}: Optimal motion planning for systems with linear differential constraints (2012).
\newblock \href {http://arxiv.org/abs/1205.5088} {\path{arXiv:1205.5088}}.

\bibitem{Zhang2022}
S.~Zhang, H.~Sang, X.~Sun, F.~Liu, Y.~Zhou, P.~Yu, A multi-objective path planning method for the wave glider in the complex marine environment, Ocean Engineering 264 (2022) 112481.
\newblock \href {https://doi.org/https://doi.org/10.1016/j.oceaneng.2022.112481} {\path{doi:https://doi.org/10.1016/j.oceaneng.2022.112481}}.

\bibitem{Garau2006AUV}
B.~Garau, A.~Alvarez, G.~Oliver, {AUV navigation through turbulent ocean environments supported by onboard H-ADCP}, in: IEEE Int. Conf. on Robotics and Automation (ICRA), 2006, pp. 3556--3561.

\bibitem{Kang2017}
W.~Kang, L.~C. Wilcox, Mitigating the curse of dimensionality: sparse grid characteristics method for optimal feedback control and {HJB} equations, Computational Optimization and Applications 68~(2) (2017) 289--315.
\newblock \href {https://doi.org/10.1007/s10589-017-9910-0} {\path{doi:10.1007/s10589-017-9910-0}}.

\bibitem{munos2002variable}
R.~Munos, A.~Moore, Variable resolution discretization in optimal control, Machine learning 49 (2002) 291--323.

\bibitem{Grune2004}
L.~Grüne, W.~Semmler, Using dynamic programming with adaptive grid scheme for optimal control problems in economics, Journal of Economic Dynamics and Control 28~(12) (2004) 2427--2456.
\newblock \href {https://doi.org/https://doi.org/10.1016/j.jedc.2003.11.002} {\path{doi:https://doi.org/10.1016/j.jedc.2003.11.002}}.

\bibitem{Festa2018}
A.~Festa, Domain decomposition based parallel howard’s algorithm, Mathematics and Computers in Simulation 147 (2018) 121--139, {APPLIED SCIENTIFIC COMPUTING XIV FOR CHALLENGING APPLICATIONS}.
\newblock \href {https://doi.org/https://doi.org/10.1016/j.matcom.2017.04.008} {\path{doi:https://doi.org/10.1016/j.matcom.2017.04.008}}.

\bibitem{Meng2024}
Y.~Meng, R.~Zhou, A.~Mukherjee, M.~Fitzsimmons, C.~Song, J.~Liu, Physics-informed neural network policy iteration: Algorithms, convergence, and verification (2024).
\newblock \href {http://arxiv.org/abs/2402.10119} {\path{arXiv:2402.10119}}.

\bibitem{Shilova2024}
A.~Shilova, T.~Delliaux, P.~Preux, B.~Raffin, {Learning HJB Viscosity Solutions with PINNs for Continuous-Time Reinforcement Learning}, Research Report (RR) 9541, Inria Lille - Nord Europe, CRIStAL - Centre de Recherche en Informatique, Signal et Automatique de Lille - UMR 9189, Univ. Lille, CNRS, Centrale Lille, Villeneuve d’Ascq, France; Univ. Grenoble Alps, CNRS, Inria, Grenoble INP, LIG, 38000 Grenoble, France (2024).

\bibitem{Mukherjee2023}
A.~Mukherjee, J.~Liu, Bridging physics-informed neural networks with reinforcement learning: {Hamilton-Jacobi-Bellman Proximal Policy Optimization (HJBPPO)} (2023).
\newblock \href {http://arxiv.org/abs/2302.00237} {\path{arXiv:2302.00237}}.

\bibitem{Barles1991}
G.~Barles, P.~E. Souganidis, Convergence of approximation schemes for fully nonlinear second order equations, Asymptotic Analysis 4~(3) (1991) 271--283.

\bibitem{butcher2008numerical}
J.~Butcher, Numerical Methods for Ordinary Differential Equations, Wiley, 2008.

\bibitem{Cruz2002}
D.~Cruz-Uribe, C.~Neugebauer, Sharp error bounds for the trapezoidal rule and simpson's rule., J. of Inequalities in Pure \& Applied Mathematics 3~(4) (2002).

\end{thebibliography}

\appendix
\section{Proof of Theorem \ref{thm:convergence}} \label{app:Proof_thm}
The proof of Theorem \ref{thm:convergence} is divided into two parts. The first one proves that the HJB equation admits a comparison principle and a unique viscosity solution, whereas the second part employs the Barles-Souganidis Theorem \cite[Theorem 2.1]{Barles1991} to show that the proposed SL numerical scheme converges to this viscosity solution.

\subsection{Viscosity solution - existence and uniqueness}

For the HJB equation in \eqref{eq:HJB_harmonic}, we can write the Hamiltonian
\begin{equation*}
    H(\xv,v,\mathbf{p}) = \sup_{\uv \in \Uset} \left\{-\ \mathbf{p} \cdot\mathbf{f}(\xv,\uv) - \left(1 - v\right)^2 \ell(\xv,\uv) \right\},
\end{equation*}
which is easily shown to be \emph{uniformly continuous in} $\xv, v$ and $\mathbf{p}$, \emph{convex in} $\mathbf{p}$ and \emph{monotone in} $v$ if $\mathbf{f}$ is Lipschitz and $\ell$ is \emph{positive semidefinite} and Lipschitz. 

In addition to this, considering that $f(.,\uv)$ and $\ell(.,\uv)$ are Lipschitz in $\xv$ (with \emph{moduli of continuity} $L_1$ and $L_2$ independent of $\uv$), it follows that (considering that the $\sup$ is attained by $\uv$ for $H(\yv,v,\mathbf{p})$)
\begin{align}
    H(\xv,v,\mathbf{p}) - H(\yv,v,\mathbf{p}) &\leq \mathbf{p} \cdot \left(\mathbf{f}(\yv,\uv) - \mathbf{f}(\xv,\uv)\right) + (1 - v)^2 \left(\ell(\yv,\uv) - \ell(\xv,\uv)\right), \notag \\
    H(\xv,v,\mathbf{p}) - H(\yv,v,\mathbf{p}) &\leq L_1 \|p\| \| \xv - \yv\| + L_2 \|\xv - \yv\|, \notag \\
    H(\xv,v,\mathbf{p}) - H(\yv,v,\mathbf{p}) &\leq L_3 \left(1 + \|p\|\right) \| \xv - \yv\|, \notag
\end{align}
with $L_3 = \max(L_1, L_2)$. Since a similar bound can be similarly found for the other difference, it follows that
\begin{align}
    |H(\xv,v,\mathbf{p}) - H(\yv,v,\mathbf{p})| &\leq L_3 \left(1 + \|p\|\right) \| \xv - \yv\|, \notag
\end{align}
and, according to \cite[Theorem 2.13]{Falcone2013}, the viscosity sub and supersolutions of \eqref{eq:HJB_harmonic} admit a \emph{comparison principle}, and, as such, equation \eqref{eq:HJB_harmonic} has a unique \emph{viscosity solution}.

\subsection{Convergence analysis}

Having shown that equation \eqref{eq:HJB_harmonic} admits a unique viscosity solution, as well as a \emph{comparison principle}, from the Barles-Souganidis Theorem \cite[Theorem 2.1]{Barles1991}, it suffices to show that the proposed numerical scheme is \emph{monotone}, a \emph{contraction mapping} and \emph{consistent} to ensure its convergence to the \emph{viscosity solution}.

\noindent $\rightarrow$\textbf{\emph{Monotonicity}}

Consider two functions $\overline{W}$ and $\overline{V}$, with $\overline{W} \leq \overline{V}$ for every point on the grid. Suppose that the $\inf$ operator in \eqref{eq:dpp__harmonic_SL} is attained by $\overline{\mathbf{w}}$ for $\overline{W}$, and $\overline{\uv}$ for $\overline{V}$. It follows that:
\begin{align}
    \overline{W}_{k}(\xv_k) &\leq \dfrac{\mathcal{I}_{\overline{W}_{k+1}}[\xv_{k+1}] + \left(1 - \mathcal{I}_{\overline{W}_{k+1}}[\xv_{k+1}]\right)g(\xv_k,\overline{\uv})}{1 + \left(1 - \mathcal{I}_{\overline{W}_{k+1}}[\xv_{k+1}]\right)g(\xv_k,\overline{\uv})}, \notag \\
    \overline{V}_{k}(\xv_k) - \overline{W}_{k}(\xv_k) &\geq \dfrac{\mathcal{I}_{\overline{V}_{k+1}}[\xv_{k+1}] + \left(1 - \mathcal{I}_{\overline{V}_{k+1}}[\xv_{k+1}]\right)g(\xv_k,\overline{\uv})}{1 + \left(1 - \mathcal{I}_{\overline{V}_{k+1}}[\xv_{k+1}]\right)g(\xv_k,\overline{\uv})} \notag\\
    &\quad - \dfrac{\mathcal{I}_{\overline{W}_{k+1}}[\xv_{k+1}] + \left(1 - \mathcal{I}_{\overline{W}_{k+1}}[\xv_{k+1}]\right)g(\xv_k,\overline{\uv})}{1 + \left(1 - \mathcal{I}_{\overline{W}_{k+1}}[\xv_{k+1}]\right)g(\xv_k,\overline{\uv})}, \notag \\
     \overline{V}_{k}(\xv_k) - \overline{W}_{k}(\xv_k) &\geq \dfrac{\mathcal{I}_{\overline{V}_{k+1}}[\xv_{k+1}] - \mathcal{I}_{\overline{W}_{k+1}}[\xv_{k+1}]}{\Omega_{\overline{V}}(\xv_k,\overline{\uv})\Omega_{\overline{W}}(\xv_k,\overline{\uv})}
\end{align}
with
\begin{align}
 \Omega_{\overline{V}}(\xv_k,\uv) &= \left(1 + \left(1 - \mathcal{I}_{\overline{V}_{k+1}}[\xv_{k+1}]\right)g(\xv_k,\uv)\right) \label{eq:OmegaV}\\
 \Omega_{\overline{W}}(\xv_k,\uv) &= \left(1 + \left(1 - \mathcal{I}_{\overline{W}_{k+1}}[\xv_{k+1}]\right)g(\xv_k,\uv)\right)
\end{align}
which implies that $\overline{V}_{k}(\xv_k) - \overline{W}_{k}(\xv_k) \geq 0$ because we employed a linear interpolation, and $\Omega_{\overline{V}}(\xv_k,\overline{\uv}) > 0$, $\Omega_{\overline{W}}(\xv_k,\overline{\uv}) > 0$, since $\overline{V} \leq 1,$ $\overline{W} \leq 1,$ $g(\xv_k,\overline{\uv}) > 0$.

\noindent $\rightarrow$\textbf{\emph{Contractiveness}}

Considering two functions $\overline{W}$ and $\overline{V}$, with $\overline{\mathbf{w}}$ minimizing $\overline{W}$. It follows that:
\begin{align}
    \overline{V}_{k}(\xv_k) - \overline{W}_{k}(\xv_k) &\leq \dfrac{\mathcal{I}_{\overline{V}_{k+1}}[\xv_{k+1}] - \mathcal{I}_{\overline{W}_{k+1}}[\xv_{k+1}]}{\Omega_{\overline{V}}(\xv_k,\overline{\mathbf{w}})\Omega_{\overline{W}}(\xv_k,\overline{\mathbf{w}})}.\notag
\end{align}
Note that, for the case in which both $\mathcal{I}_{\overline{V}_{k+1}}[\xv_{k+1}]$ and $\mathcal{I}_{\overline{W}_{k+1}}[\xv_{k+1}]$ are equal to 1, the inequality is trivially upper bounded by anything larger than zero. As such, we can consider, as a worst-case bound, the case in which one of them is one and the other is $1 - \varepsilon$, leading to
\begin{align}
     \overline{V}_{k}(\xv_k) - \overline{W}_{k}(\xv_k) &\leq \dfrac{1}{1 + \varepsilon g(\xv_k,\overline{\mathbf{w}})} \|\overline{V} - \overline{W}\|_\infty^{(k+1)}, \notag
\end{align}
with $\|\overline{V} - \overline{W}\|_\infty^{(k+1)}$ being the maximum of the error between the two functions in the next time step. Since we are assuming that $\ell(\xv,\uv) > 0$, then $g(\xv_k,\uv_k) > \overline{g}\Delta t > 0 ~~ \forall \xv_k, \uv_k$, and a similar bound can be found for $\overline{W}_{k}(\xv_k) - \overline{V}_{k}(\xv_k)$, then:
\begin{align}
    \|\overline{V} - \overline{W}\|_\infty^{(k)} &\leq \dfrac{1}{1 + \varepsilon \overline{g}\Delta t} \|\overline{V} - \overline{W}\|_\infty^{(k+1)}. \label{eq:contractRateHarmonic}
\end{align}
As we solve the problem back in time, this shows that our approximation scheme is a contraction mapping. From the Banach fixed-point Theorem, it guarantees that our approximation scheme converges to a unique solution.

\noindent $\rightarrow$\textbf{\emph{Consistency}} 

We start our consistency analysis by considering the error of time discretization, comparing solutions from \eqref{eq:DPP_harmonic} and \eqref{eq:dpp__harmonic_discrete}. If we consider that the $\inf$ operator is attained by $\overline{\uv}$ in \eqref{eq:dpp__harmonic_discrete}, it follows that:
\begin{align*}
    &\bar{v}(\xv(t)) - \bar{v}_k(\xv_{k}) \leq \notag \\
    &\quad  \dfrac{\bar{v}(\yv_x(\Delta t, \overline{\uv})) + \left(1 - \bar{v}(\yv_x(\Delta t, \overline{\uv}))\right)\int_0^{\Delta t}\ell(\yv_x(t, \overline{\uv}),\overline{\uv}) dt}{1 + \left(1 - \bar{v}(\yv_x(\Delta t, \overline{\uv}))\right)\int_0^{\Delta t}\ell(\yv_x(t, \overline{\uv}),\overline{\uv}) dt} \notag \\ 
    &\quad-\dfrac{\bar{v}_{k}(\xv_{k+1}) + \left(1 - \bar{v}_{k}(\xv_{k+1})\right)g(\xv_k,\overline{\uv})}{1 + \left(1 - \bar{v}_{k}(\xv_{k+1})\right)g(\xv_k,\overline{\uv})}.
\end{align*}

Some algebraic manipulations lead to:
\begin{align}
    &\bar{v}(\xv(t)) - \bar{v}_k(\xv_{k}) \leq \notag \\
    &\quad \dfrac{\left(\bar{v}(\yv_x(\Delta t, \overline{\uv})) - \bar{v}_{k}(\xv_{k+1})\right)}{\left(1 + q(\xv(t),\overline{\uv})\right)\left(1 + \left(1 - \bar{v}_{k}(\xv_{k+1})\right)g(\xv_k,\overline{\uv})\right)} \notag \\
    &\quad + \dfrac{\left(1 - \bar{v}(\yv_x(\Delta t, \overline{\uv}))\right)\left(1 - \bar{v}_{k}(\xv_{k+1})\right) \Delta g}{\left(1 + q(\xv(t),\overline{\uv})\right)\left(1 + \left(1 - \bar{v}_{k}(\xv_{k+1})\right)g(\xv_k,\overline{\uv})\right)}, \label{eq:cons_init_step} \\
    &q(\xv(t),\overline{\uv}) = \left(1 - \bar{v}(\yv_x(\Delta t, \overline{\uv}))\right)\int_0^{\Delta t}\ell(\yv_x(t, \overline{\uv}),\overline{\uv}) dt, \\
    &\Delta g = \int_0^{\Delta t}\ell(\yv_x(t, \overline{\uv}),\overline{\uv}) dt - g(\xv_k,\overline{\uv}).
\end{align}

From \cite[Lemma 406B]{butcher2008numerical}, we know that, using a trapezoidal method to solve the system dynamics in \eqref{eq:sys_dynamics} leads to a bound (over one time step)
\begin{align*}
    \|\yv_x(\Delta t, \overline{\uv}) - \xv_{k+1}\| \leq LM\Delta t^2,
\end{align*}
with $L$ the Lipschitz constant of $\mathbf{f}(\xv,\uv)$ with regards to $\xv$, and $\|\mathbf{f}(\xv,\uv)\| \leq M$. Note that tighter bounds are available for the local truncation error of the trapezoidal method (the error over one time step), but these usually require extra assumptions about the smoothness of $\mathbf{f}$. In addition to this, from \cite[Corollary 1.4]{Cruz2002}, considering that $\ell(t)$ is Lipschitz, we have that
\rev{
\begin{align*}
    \left|\int_0^{\Delta t}\!\!\!\!\!\ell(\yv_x(t, \overline{\uv})dt - g(\xv_k,\overline{\uv}) \right|
    &\leq \left|\int_0^{\Delta t}\!\!\!\!\!\ell(\yv_x(t, \overline{\uv})dt - \dfrac{\left(\ell(\xv_k,\bar{\uv}) + \ell(\yv_x(\Delta t, \bar{\uv}),\bar{\uv}) \right)}{2}\right| \notag \\
    &\quad + \left|\dfrac{\left(\ell(\xv_{k+1},\bar{\uv}) - \ell(\yv_x(\Delta t, \bar{\uv}),\bar{\uv}) \right)}{2}\right|, \notag \\ 
    & \leq \dfrac{\Delta t^2}{8}\left(\sup \dot{\ell} - \inf \dot{\ell}\right) + KLM\Delta t^2, 
\end{align*}
}
with $K$ being the Lipschitz constant of $\ell$ with regards to $\xv$, and $C_1$ some constant. 

If we consider that, $\bar{v}$ is Lipschitz continuous, it follows that
\begin{align}
    \left|\int_0^{\Delta t} \ell(\yv_x(t, \bar{\uv}),\bar{\uv})dt - g(\xv_k,\bar{\uv})\right| &\leq C_1 \Delta t^2,  \\
    \left|\bar{v}(\yv_x(\Delta t, \overline{\uv})) - \bar{v}(\xv_{k+1})\right| &\leq C_2 \Delta t^2,
\end{align}
and that the worst case upper bound for \eqref{eq:cons_init_step} happens when one of the value functions is equal to 1, while the other is equal to $1-\varepsilon$, it follows that
\begin{align}
    \bar{v}(\xv(t)) - \bar{v}_k(\xv_{k}) \leq& \dfrac{C_3 \Delta t^2 + \|\bar{v} - \bar{v}_k\|_\infty}{1 + \varepsilon \bar{g} \Delta t}. 
    \label{eq:bound1_harmonic}
\end{align}

If we consider that the $\inf$ operator is attained by $\uv^\ast$ in~\eqref{eq:DPP_harmonic}, and that 
\begin{align}
    \hat{\uv}_k = \frac{1}{\Delta t} \int_0^{\Delta t} \uv^\ast(\tau) d\tau \label{eq:mean_control}
\end{align}
is the control obtained by the mean of the optimal control over a time step, it follows that
\begin{align}
    \bar{v}_k(\xv_{k}) - \bar{v}(\xv(t)) \leq&  \dfrac{\bar{v}_{k}(\xv_{k+1}) + \left(1 - \bar{v}_{k}(\xv_{k+1})\right)g(\xv_k,\hat{\uv}_k)}{1 + \left(1 - \bar{v}_{k}(\xv_{k+1})\right)g(\xv_k,\hat{\uv}_k)} \notag \\
    &-\dfrac{\bar{v}(\yv_x(\Delta t, \uv^\ast)) + \left(1 - \bar{v}(\yv_x(\Delta t, \uv^\ast))\right)\int_0^{\Delta t}\ell(\yv_x(t, \uv^\ast),\uv^\ast) dt}{1 + \left(1 - \bar{v}(\yv_x(\Delta t, \uv^\ast))\right)\int_0^{\Delta t}\ell(\yv_x(t, \uv^\ast),\uv^\ast) dt}. \notag
\end{align}

Some algebraic manipulations lead to:
\begin{align}
    &\bar{v}_k(\xv_{k}) - \bar{v}(\xv(t)) \leq \notag \\
    &\; \dfrac{\left(\bar{v}_{k}(\xv_{k+1}) - \bar{v}(\yv_x(\Delta t, \uv^\ast))\right)}{\left(1 + \left(1 - \bar{v}_{k}(\xv_{k+1})\right)g(\xv_k,\hat{\uv}_k)\right)\left(1 + \bar{q}(\xv(t),\uv^\ast)\right)} \notag \\
    &\;  + \dfrac{\left(1 - \bar{v}(\yv_x(\Delta t, \uv^\ast))\right)\left(1 - \bar{v}_{k}(\xv_{k+1})\right) \Delta \hat{g}}{\left(1 + \left(1 - \bar{v}_{k}(\xv_{k+1})\right)g(\xv_k,\hat{\uv}_k)\right)\left(1 + \bar{q}(\xv(t),\uv^\ast)\right)}, \label{eq:const_inter_step} \\
    &\bar{q}(\xv(t),\uv^\ast) = \left(1 - \bar{v}(\yv_x(\Delta t, \uv^\ast))\right)\int_0^{\Delta t}\ell(\yv_x(t, \uv^\ast),\uv^\ast) dt, \\
    &\Delta \hat{g} = \left(g(\xv_k,\hat{\uv}_k) - \int_0^{\Delta t}\ell(\yv_x(t, \uv^\ast),\uv^\ast) dt\right).\notag
\end{align}

Consider that
\begin{align*}
    \yv_x(\Delta t, \uv^\ast) &= \xv_k + \int_0^{\Delta t} \mathbf{f}(\yv_x(\tau, \uv^\ast(\tau)) d\tau, \\
    \xv_{k+1} &= \xv_k + \frac{\Delta t}{2}\left(\mathbf{f}(\xv_k,\hat{\uv}_k) + \mathbf{f}(\xv_{k+1},\hat{\uv}_k)\right),
\end{align*}
since $\mathbf{f}$ can be decomposed, as in \eqref{eq:sys_affine}, it follows that 
\rev{
\begin{align}
    \yv_x(\Delta t, \uv^\ast) &= \xv_k + \int_0^{\Delta t} \mathbf{f}_1(\yv_x(\tau, \uv^\ast(\tau))d\tau \notag \\
    &\quad +\int_0^{\Delta t} \dfrac{\left(F_2(\yv_x(\tau, \uv^\ast(\tau)) - F_2(\xv_k)\right)}{2}\uv^\ast(\tau)d\tau \notag \\
    &\quad +\int_0^{\Delta t} \dfrac{\left(F_2(\yv_x(\tau, \uv^\ast(\tau)) - F_2(\xv_{k+1})\right)}{2}\uv^\ast(\tau)d\tau \notag \\
    &\quad +\dfrac{F_2(\xv_k)}{2}\int_0^{\Delta t} \uv^\ast(\tau)d\tau +\dfrac{F_2(\xv_{k+1})}{2}\int_0^{\Delta t} \uv^\ast(\tau)d\tau, \notag
\end{align}
}
\rev{
\begin{align}
    \yv_x(\Delta t, \uv^\ast) &= \xv_k + \int_0^{\Delta t} \mathbf{f}_1(\yv_x(\tau, \uv^\ast(\tau))d\tau \notag \\
    &\quad +\int_0^{\Delta t} \dfrac{\left(F_2(\yv_x(\tau, \uv^\ast(\tau)) - F_2(\xv_k)\right)}{2}\uv^\ast(\tau)d\tau \notag \\
    &\quad +\int_0^{\Delta t} \dfrac{\left(F_2(\yv_x(\tau, \uv^\ast(\tau)) - F_2(\xv_{k+1})\right)}{2}\uv^\ast(\tau)d\tau \notag \\
    &\quad +\dfrac{F_2(\xv_k)\Delta t}{2}\hat{\uv}_k + \dfrac{F_2(\xv_{k+1})\Delta t}{2}\hat{\uv}_k, \notag
\end{align}
}
\rev{
\begin{align}
    \yv_x(\Delta t, \uv^\ast) - \xv_{k+1} &= \int_0^{\frac{\Delta t}{2}} \left(\mathbf{f}_1(\yv_x(\tau, \uv^\ast(\tau)) - \mathbf{f}_1(\xv_k)\right) d\tau \notag \\
    &\quad + \int_{\frac{\Delta t}{2}}^{\Delta t} \left(\mathbf{f}_1(\yv_x(\tau, \uv^\ast(\tau)) - \mathbf{f}_1(\xv_{k+1})\right) d\tau \notag \\
    &\quad +\int_0^{\Delta t} \dfrac{\left(F_2(\yv_x(\tau, \uv^\ast(\tau)) - F_2(\xv_k)\right)}{2}\uv^\ast(\tau)d\tau \notag \\
    &\quad +\int_0^{\Delta t} \dfrac{\left(F_2(\yv_x(\tau, \uv^\ast(\tau)) - F_2(\xv_{k+1})\right)}{2}\uv^\ast(\tau)d\tau. \notag
\end{align}
}

If we consider that  $\mathbf{f}_1$ and $F_2$ are Lipschitz in $\xv$, and that $\uv$ is bounded, it follows that
\begin{align}
    &\|\yv_x(\Delta t, \uv^\ast) - \xv_{k+1}\| \leq L_1 \int_0^{\frac{\Delta t}{2}} \|\yv_x(\tau, \uv^\ast(\tau)) - \xv_k\| d\tau \notag \\ &+ L_1 \int_{\frac{\Delta t}{2}}^{\Delta t} \|\yv_x(\tau, \uv^\ast(\tau)) - \xv_{k+1})\| d\tau \notag \\
    &+ L_2 \int_0^{\Delta t} \|\yv_x(\tau, \uv^\ast(\tau)) - \xv_k\| d\tau \notag \\ &+ L_2 \int_0^{\Delta t} \|\yv_x(\tau, \uv^\ast(\tau)) - \xv_{k+1})\| d\tau, \notag
\end{align}
\begin{align}
    &\|\yv_x(\Delta t, \uv^\ast) - \xv_{k+1}\| \leq L_1 \int_0^{\frac{\Delta t}{2}} \|\yv_x(\tau, \uv^\ast(\tau)) - \xv_k\| d\tau \notag \\&+ L_1 \int_{0}^{\Delta t} \|\yv_x(\tau, \uv^\ast(\tau)) - \xv_{k+1})\| d\tau \notag \\
    &+ L_2 \int_0^{\Delta t} \|\yv_x(\tau, \uv^\ast(\tau)) - \xv_k\| d\tau \notag\\&+ L_2 \int_0^{\Delta t} \|\yv_x(\tau, \uv^\ast(\tau)) - \xv_{k+1})\| d\tau. \notag
\end{align}

Considering that $\|\mathbf{f}\| \leq M, \forall \xv,\uv$, it follows that
\begin{align}
    \|\yv_x(\tau, \uv^\ast) - \xv_k\| \leq M \tau, \notag
\end{align}
and we arrive at the bound
\rev{
\begin{align}
    \|\yv_x(\Delta t, \uv^\ast) - \xv_{k+1}\| &\leq (L_1 + L_2) \int_{0}^{\Delta t} \|\yv_x(\tau, \uv^\ast) - \xv_{k+1}\| d\tau \notag \\
    &\quad + \frac{(L_1 + L_2) M\Delta t^2}{8}, \notag \\
    \|\yv_x(\Delta t, \uv^\ast) - \xv_{k+1}\| &\leq \bar{L} \int_{0}^{\Delta t} \|\yv_x(\tau, \uv^\ast) - \xv_{k+1}\| d\tau + \bar{M} \Delta t^2, \notag
\end{align}
}
which, from the integral form of the Gronwall-Bellman inequality leads to
\begin{align}
    \|\yv_x(\Delta t, \uv^\ast) - \xv_{k+1}\| \leq \bar{M} \Delta t^2 e^{\bar{L}\Delta t}. \notag
\end{align}

Considering that
\rev{
\begin{align}
    g(\xv_k,\hat{\uv}_k) \!-\! \int_0^{\Delta t}\!\!\!\!\ell(\yv_x(t, \uv^\ast),\uv^\ast) dt &= \int_0^{\frac{\Delta t}{2}} \!\!\!\!\!\Big(\ell(\xv_k,\hat{\uv}) \textcolor{\replycolor}{ - \ell(\xv_k,\uv^\ast) + \ell(\xv_k,\uv^\ast)} \notag \\
    &\quad\quad\quad - \ell(\yv_x(t, \uv^\ast),\uv^\ast) \Big) dt \notag \\
    &\; + \int_{\frac{\Delta t}{2}}^{\Delta t} \!\!\!\!\Big(\ell(\xv_{k+1},\hat{\uv}) \textcolor{\replycolor}{- \ell(\yv_x(\Delta t, \uv^\ast),\uv^\ast)}\Big) dt  \notag \\
    &+ \int_{\frac{\Delta t}{2}}^{\Delta t} \!\!\!\!\Big(\textcolor{\replycolor}{\ell(\yv_x(\Delta t, \uv^\ast),\uv^\ast)} - \ell(\yv_x(t, \uv^\ast),\uv^\ast)\Big) dt, \notag
\end{align}
}
and, considering that $\ell$ is convex in $\uv$, from Jensen's inequality, it follows that
\begin{align}
    &-\int_0^{\frac{\Delta t}{2}} \ell(\xv_k,\uv^\ast) dt \leq - \frac{\Delta t}{2} \ell(\xv_k,\hat{\uv}_k) \\
    &-\int_{\frac{\Delta t}{2}}^{\Delta t} \ell(\yv_x(\Delta t, \uv^\ast),\uv^\ast) dt \leq - \frac{\Delta t}{2} \ell(\yv_x(\Delta t, \uv^\ast),\hat{\uv}_k), \notag
\end{align}
so that
\begin{align}
    &g(\xv_k,\hat{\uv}_k) - \int_0^{\Delta t}\ell(\yv_x(t, \uv^\ast),\uv^\ast) dt \leq \notag\\
    &\quad\int_0^{\frac{\Delta t}{2}} \left(\ell(\xv_k,\uv^\ast) - \ell(\yv_x(t, \uv^\ast),\uv^\ast) \right) dt \notag \\
    &\quad + \int_{\frac{\Delta t}{2}}^{\Delta t} \left(\ell(\xv_{k+1},\hat{\uv}_k) - \ell(\yv_x(\Delta t, \uv^\ast),\hat{\uv}_k)\right)dt \notag \\ &\quad+ \int_{\frac{\Delta t}{2}}^{\Delta t} \left(\ell(\yv_x(\Delta t, \uv^\ast),\uv^\ast) - \ell(\yv_x(t, \uv^\ast),\uv^\ast)\right) dt. \notag
\end{align}

Considering that $\ell$ has Lipschitz constant $K$ with respect to $\xv$, it follows that
\rev{
\begin{align}
    g(\xv_k,\hat{\uv}_k) - \int_0^{\Delta t}\!\!\!\!\ell(\yv_x(t, \uv^\ast),\uv^\ast) dt &\leq \int_0^{\frac{\Delta t}{2}} \!\!\!\! KM\tau d\tau \notag \\
    &\quad + \int_{\frac{\Delta t}{2}}^{\Delta t} \!\!\!\! \left(K\bar{M} \Delta t^2 e^{\bar{L}\Delta t} + KM\tau \right) d\tau, \notag\\
    g(\xv_k,\hat{\uv}_k) - \int_0^{\Delta t}\ell(\yv_x(t, \uv^\ast),\uv^\ast) dt &\leq \frac{KM\Delta t^2}{2} + \frac{K \bar{M} \Delta t^3}{2} e^{\bar{L}\Delta t}. \notag
\end{align}
}

If we consider that, $\bar{v}$ is Lipschitz continuous, it follows that
\begin{align}
    g(\xv_k,\hat{\uv}_k) - \int_0^{\Delta t}\ell(\yv_x(t, \uv^\ast),\uv^\ast) dt &\leq C_4 \Delta t^2 + C_5 \Delta t^3 e^{\bar{L} \Delta t} \notag \\
    |\bar{v}(\xv_{k+1}) - \bar{v}(\yv_x(\Delta t, \uv^\ast))| &\leq C_6 \Delta t^2 e^{\bar{L} \Delta t}. \notag
\end{align}
Considering also that the worst case upper bound in \eqref{eq:const_inter_step} happens when one of the value functions is equal to 1, while the other is equal to $1-\varepsilon$, it follows that
\begin{align}
    \bar{v}_k(\xv_{k}) - \bar{v}(\xv(t)) \leq \dfrac{C_4 \Delta t^2 + \left(C_5 \Delta t^3 + C_6 \Delta t^2\right)e^{\bar{L} \Delta t} + \|\bar{v} - \bar{v}_k\|_\infty}{1 + \varepsilon \bar{g} \Delta t}. \label{eq:bound2_harmonic}
\end{align}

By combining \eqref{eq:bound1_harmonic} and \eqref{eq:bound2_harmonic}, we have that
\begin{align}
    \|\bar{v} - \bar{v}_k\|_\infty &\leq \dfrac{C_7 \Delta t^2 + \left(C_5 \Delta t^3 + C_6 \Delta t^2\right)e^{\bar{L} \Delta t}}{\varepsilon \bar{g} \Delta t} \\
    &\approx \dfrac{C_7 \Delta t^2 + \left(C_5 \Delta t^3 + C_6 \Delta t^2\right)(1 + \bar{L} \Delta t)}{\varepsilon \bar{g} \Delta t}, \notag
\end{align}
in which the last approximation comes from considering that, for $\theta$ close to zero, $e^\theta \approx 1 + \theta$. Asymptotically, this bound is \emph{dominated} by the linear term when $\Delta t \rightarrow 0$, since the higher order terms vanish faster. In that regard, we will write
\begin{align}
    \|\bar{v} - \bar{v}_k\|_\infty \leq C \Delta t. \label{eq:error_time_harmonic}
\end{align}

For the space discretization error, we analyze the errors of the value function on the grid points, by comparing \eqref{eq:dpp__harmonic_discrete} and \eqref{eq:dpp__harmonic_SL}. To differentiate them, we will denote the value on the grid points of \eqref{eq:dpp__harmonic_SL} by $\overline{V}_k$. If we consider that the $\inf$ is attained by $\overline{\uv}$ in \eqref{eq:dpp__harmonic_discrete}, it follows that:
\begin{align}
    \overline{V}_k(\xv_k) - \bar{v}_k(\xv_k) \leq& \dfrac{\mathcal{I}_{\bar{V}_{k+1}}[\xv_{k+1}] - \bar{v}_{k+1}(\xv_{k+1})}{\Omega_{\overline{V}}(\xv_k,\overline{\uv})\Omega_{\bar{v}}(\xv_k,\overline{\uv})}, \notag
\end{align}
with $\Omega_{\overline{V}}$ from \eqref{eq:OmegaV} and
\begin{align}
    \Omega_{\bar{v}}(\xv_k,\overline{\uv}) = 1 + \left(1 - \bar{v}_{k+1}(\xv_{k+1})\right)g(\xv_k,\overline{\uv}). \notag
\end{align}
By considering that $\overline{V}$ is Lipschitz, $g(\xv_k,\overline{\uv}) > \overline{g}\Delta t$, and that the worst case upper bound happens when one of the value functions is equal to 1, while the other is equal to $1-\varepsilon$ leads to:
\begin{align}
    \overline{V}_k(\xv_k) - \bar{v}_k(\xv_k) \leq& \dfrac{C_8 \Delta x + \|\overline{V_k} - \overline{v}_k\|_\infty}{1 + \varepsilon \overline{g} \Delta t}. \notag
\end{align}
with $C_8$ being a constant and $\Delta x$ being the largest distance between any point and a grid point.

Since a similar bound can be found for $\bar{v}_k(\xv_k) - \overline{V}_k(\xv_k)$, then:
\begin{align}
     \|\overline{V_k} - \overline{v}_k\|_\infty \leq& \dfrac{C_8 \Delta x + \|\overline{V_k} - \overline{v}_k\|_\infty}{1 + \varepsilon \overline{g} \Delta t}\notag \\
     \|\overline{V_k} - \overline{v}_k\|_\infty \leq& C_9 \dfrac{\Delta x}{\Delta t}. \label{eq:error_space_harmonic}
\end{align}
with $C_9$ a constant. Combining \eqref{eq:error_time_harmonic} and \eqref{eq:error_space_harmonic}, we have that:
\begin{align}
    \|\overline{v} - \overline{V}\|_\infty \leq C_{10} \Delta t + C_{9} \frac{\Delta x}{\Delta t},
    \label{eq:consistency_harmonic}
\end{align}
with $C_{10}$ a constant. Similarly to the original Kruzkov transformation, the best coupling between $\Delta t$ and $\Delta x$, in this case, is given by $\Delta x = \Delta t^2$, indicating that \textbf{our grid resolution should be finer than our time discretization resolution}.

Since we have shown that our scheme is \textbf{monotone, a contraction mapping, and consistent}, from the Barles-Souganidis Theorem, we have proven that it is \textbf{convergent}. $\qed$

\end{document}